\documentclass[lettersize,journal]{IEEEtran}
\usepackage{amsmath,amsfonts}
\usepackage{algorithmic}
\usepackage{algorithm}
\usepackage{array}
\usepackage[caption=false,font=normalsize,labelfont=sf,textfont=sf]{subfig}
\usepackage{textcomp}
\usepackage{stfloats}
\usepackage{url}
\usepackage{verbatim}
\usepackage{graphicx}
\usepackage{cite}
\usepackage{algorithm}
\usepackage{algorithmic}
\usepackage{threeparttable}
\usepackage{amsmath,amsfonts}
\usepackage{color,xcolor}
\usepackage{colortbl}
\usepackage{soul}
\usepackage{threeparttable}
\begin{document}

\title{Neighborhood Contrastive Transformer for \\ Change Captioning}

\author{Yunbin Tu, Liang Li, Li Su, Ke Lu, and  Qingming Huang,~\IEEEmembership{Fellow,~IEEE}

\thanks{The work was supported by the National Key R\&D Program of China under Grant 2018AAA0102000, and in part by the National Natural Science Foundation of China: U21B2038, 61931008, and by Youth Innovation Promotion Association of CAS under Grant 2020108.
(\emph{Corresponding author: Liang Li, Li Su.})}

\thanks{Yunbin Tu is with the School of Computer Science and Technology, University of Chinese Academy of Sciences, Beijing 101408, China (e-mail: tuyunbin22@mails.ucas.ac.cn).}
\thanks{Liang Li is with Institute of Computing Technology, Chinese Academy of Sciences, Beijing 100190, China (e-mail: liang.li@ict.ac.cn)}
\thanks{Li Su is with the School of Computer Science and
Technology, University of Chinese Academy of Sciences, Beijing 101408, China (e-mail: Suli@ucas.ac.cn).}
\thanks{Ke Lu is with the School of Engineering Science, University of Chinese Academy of Sciences, Beijing, China, with Peng Cheng Laboratory, Nanshan, Shenzhen, Guangdong, China. (e-mail: luk@ucas.ac.cn).}

\thanks{Qingming Huang is with the School of Computer Science and Technology,
University of Chinese Academy of Sciences, Beijing 101408, China,
with Institute of Computing Technology, Chinese Academy of Sciences,
Beijing 100190, China (e-mail: qmhuang@ucas.ac.cn).}

\thanks{This paper has supplementary downloadable material available at http://ieeexplore.ieee.org., provided by the author. The material includes the implementation details and more qualitative examples on the three datasets. This material is 0.98 MB in size.}}

\markboth{Journal of \LaTeX\ Class Files,~Vol.~14, No.~8, August~2021}%
{Shell \MakeLowercase{\textit{et al.}}: A Sample Article Using IEEEtran.cls for IEEE Journals}


\maketitle

\begin{abstract}
Change captioning is to describe the semantic change between a pair of similar images in natural language.  It is more challenging than general image captioning, because it requires capturing fine-grained change information while being immune to irrelevant viewpoint changes, and solving syntax ambiguity in change descriptions.  In this paper, we propose a neighborhood contrastive transformer to improve the model's perceiving ability for various changes under different scenes and cognition ability for complex syntax structure. Concretely, we first design a neighboring feature aggregating to integrate neighboring context into each feature, which helps quickly locate the inconspicuous changes under the guidance of conspicuous referents. Then, we devise a common feature distilling to compare two images at neighborhood level and extract common properties from each image, so as to learn effective contrastive information between them. Finally, we introduce the explicit dependencies between words to calibrate the transformer decoder, which helps better understand complex syntax structure during training. Extensive experimental results demonstrate that the proposed method achieves the state-of-the-art performance on three public datasets with different change scenarios. The code is available at \url{https://github.com/tuyunbin/NCT}.
\end{abstract}


\section{Introduction}
\IEEEPARstart{C}{hange} captioning aims to describe what has changed between two semantically similar images, which is a novel task in the community of vision and language \cite{9488296,9780231,9807440}. It extends the conventional image captioning \cite{9146724,9468925} further, \emph{i.e.},  it needs to simultaneously deal with two images and describe their disagreement. This pushes forward the research of exploring the relationship and difference of image pair.  
In addition, it has wide applications, such as providing explanation of complex image editing effects for laypersons or visually-impaired users, outputting logs about monitored areas, and generating reports about pathological changes \cite{jhamtani2018learning,tan2019expressing,shi2020finding}.

\begin{figure}[t]
\centering
\includegraphics[width=0.45\textwidth]{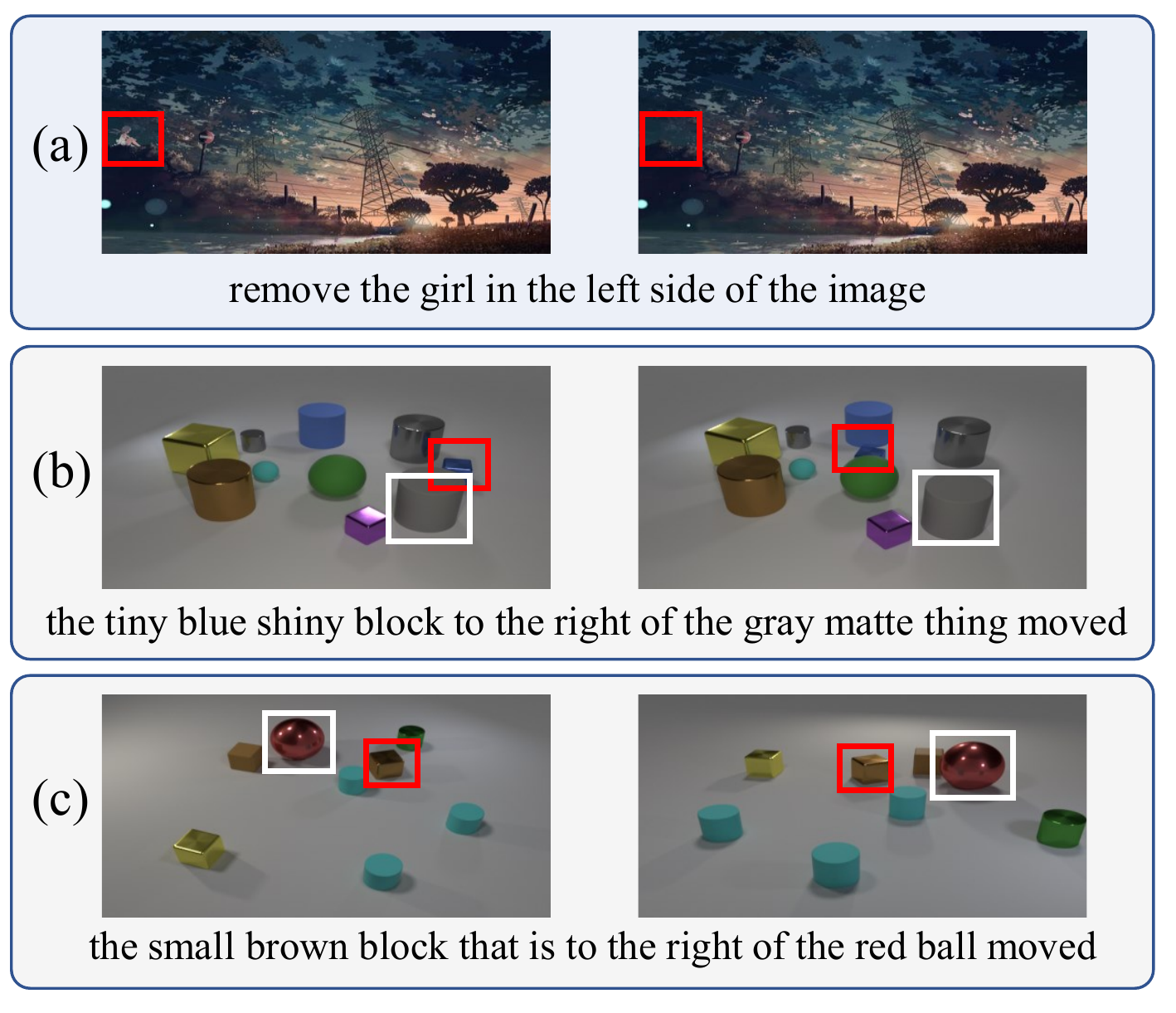} 
\caption{The examples of change captioning. The first one is from image editing scene, where the removed object is inconspicuous. The second one shows that with both object move and moderate viewpoint change, and the changed object is partially occluded. The last one shows that with both object move and extreme viewpoint change, where the real movement is overwhelmed by pseudo movements. The changed objects and referents are shown in the red and white boxes, respectively} %
\label{fig1}
\end{figure}

The key challenges are mainly embodied in the two aspects. First, the model should have the ability of fine-grained semantic comprehension, because change information is usually hard to pinpoint. For example, in Fig. \ref{fig1} (a), the removed girl is easy to ignore, due to her inconspicuous position and vague shape. In Fig. \ref{fig1} (b), The change is hard to be located, because the moved block is partially occluded. Second, the model should be immune to irrelevant distractors and only describe genuine semantic change. In a dynamic environment, it is 
nearly impossible to acquire two images under same viewpoint due to various factors, such as camera shaking, different shoot time, etc. 
In Fig. \ref{fig1} (c), extreme viewpoint change leads to obvious pseudo movement for unchanged objects, which could overwhelm the real change and mislead the model into generating inaccurate sentences.

There have been previous endeavors for the above challenges. Despite progresses, these methods suffer from learning the effective change representation. Specifically, they compare two images mainly at global or local level, where global refers to direct subtraction \cite{park2019robust,hosseinzadeh2021image} and local is to compute their similarity based on individual feature matching \cite{shi2020finding,kim2021agnostic,yao2022image}. The former is too coarse to capture inconspicuous or occluded changes. The latter is more reasonable, but it is easily influenced by extreme viewpoint change, \emph{e.g.,} in Fig. \ref{fig1} (c) every object seems to move. In this case, such individual matching is unable to learn the stable features of change. We argue that to learn effective features across viewpoint changes, the model should compare details at neighborhood level. The reasons are that spatially neighboring objects are highly correlated in an image \cite{wu2021cvt}, where if an object changed, its relations with neighboring objects  would change as well. Such relation change helps mine those inconspicuous  changes. Besides, pseudo changes are actually the distortion of objects' scale and location, 
so the relations of these neighboring objects are not affected. Considering this, we try to dynamically integrate features at neighborhood level, thus helping the model resist viewpoint changes and locate the real change.

Besides, we observe that a change description usually consists of two parts: the semantic change and a referent, which makes it contain complex syntax structure. As shown in Fig. \ref{fig1} (c),  the main clause of this sentence is ``the small brown block moved''. However, the subject ``block'' and its predicate ``moved'' are separated by a subordinate clause describing the referent ``ball''. In this case, the word ``moved'' is closer to the word of ``ball'' than ``block''. During training, if a model does not understand syntax relations between words, it might learn wrong information from the ground-truth caption. To the best of our knowledge, this problem is disregarded by the existing methods. In fact, the above misunderstanding could be avoided if the model notices the direct dependency relation between ``block'' and ``moved''. Hence, it is necessary to introduce explicit dependency relations during training, which helps the model understand the syntax structure of captions.

In this paper, we propose a Neighborhood Contrastive Transformer to pinpoint change under different change scenarios, and endow the model with the syntax knowledge of dependency relation to address structural ambiguity. Concretely, given an image pair, a neighborhood feature aggregating is first designed to integrate neighboring context into features of each image. This helps the model resist viewpoint changes and perceive the fine-grained change under the guidance of neighboring referents. Then, based on similarity matching, a common feature distilling is customized to establish correspondences between the above two image features, so as to summarize their common features.  Next, the stable features of change in each image are computed by removing common features, which are fused to learn contrastive features between the image pair. These contrastive features are subsequently fed into a transformer decoder to generate descriptions. During training, we provide the decoder with the prior knowledge of dependencies between words, which is beneficial to understand the complex syntax structure in ground-truth captions.

The contributions of this paper are summarized below: (1) A neighborhood contrastive transformer is proposed to pinpoint changes via performing neighborhood contrast between image pairs, where a neighborhood feature aggregating is designed to explore fine-grained changes and resist viewpoint change; a common feature distilling is devised to capture discriminative properties of each image and construct their contrastive features for sentence generation. (2) This work is the first attempt in this task to solve syntax structural ambiguity via introducing explicit dependencies between words. (3) Extensive experiments demonstrate that our method performs favorably against the state-of-the-art methods on three public datasets.

\section{Related Work}
\label{related work}
\textbf{Image/Video Captioning.} Before introducing the works of change captioning, we first review recently published works in conventional image/video captioning. TTA \cite{tu2021enhancing}  detects visual tags from videos to bridge visual-textual gap, and presents a textual-temporal attention model to build alignment between words and frames. LSRT \cite{9741388} proposes the long short-term relation transformer to fully mine objects' relations for caption generation. I$^2$Transformer \cite{9738841} learns the intra- and inter-relation embedded representation from different modalities, which is fed into the standard transformer for caption generation. P+D attention \cite{9400759} proposes a dual attention module on pyramid image feature maps, which can explore the visual-semantic correlations and refine generated captions. MSA \cite{9761944} presents a multi-branch self-attention and duplicates it multiple times, in order to increase the expressive power of general self-attention model during caption generation.  HTG+HMG \cite{tu2023relation} proposes a relation-aware attention  by designing two kinds of graphs, namely  linguistics-to-vision heterogeneous graph  and vision-to-vision homogeneous graph.
 
\textbf{Change Captioning.} It is a new task in visual captioning, while it is more challenging. This is because it needs to understand the contents of two images, and further to describe their difference. 
 The pioneer work  \cite{jhamtani2018learning}  describes  the change based on the surveillance scenarios. The work  \cite{tan2019expressing}  elaborates the editing transformation between two images, as shown in Fig.1 (a). The common point of these two works is that they detect and describe changes between two well-aligned images. In fact, there exist viewpoint shifts as we shoot pictures, which poses a challenge to distinguish the real change from pseudo changes.  Considering this, Park \emph{et al.} \cite{park2019robust} and Kim \emph{et al.}\cite{kim2021agnostic} respectively release two datasets with moderate (Fig.\ref{fig1} (b)) and extreme viewpoint changes (Fig.\ref{fig1} (c)). To describe semantic change under viewpoint changes, Park \emph{et al.} propose  a DUDA model for localizing and describing changes, where they model the difference by subtracting two unaligned images, which might compute the difference features with  noise \cite{tu2021semantic}.

To ease this problem,  Hosseinzadeh \emph{et al.} \cite{hosseinzadeh2021image} leverage a retrieval model of TIRG \cite{vo2019composing} to regularize DUDA. Tu \emph{et al.} \cite{tu2021semantic} measure the relations between the subtracted change and image pair to judge if the change has actually happened.  Instead of using direct subtraction, on the one hand, the works  \cite{shi2020finding,kim2021agnostic,huang2021image}  first distill the common features between two images based on feature similarity. Then, they remove these features to explicitly capture the features of change. On the other hand, the works  \cite{Qiu_2021_ICCV,yao2022image}  match the similar features between two images to implicitly infer the features of change. To enhance the visual-textual alignment, Kim \emph{et al.} \cite{kim2021agnostic} introduce a  cycle consistency module to refine generated sentences. Yao \emph{et al.} \cite{yao2022image} model the fine-grained cross-modal alignment by the paradigm of pre-training to fine-tuning.  

In addition, Liao \emph{et al.} \cite{liao2021scene}  introduce the 3D information of depths of objects to deal with viewpoint changes. They first input images into a pre-trained depth estimation model to obtain the depth maps. Meanwhile, they utilize a pre-trained Yolov4 to obtain the bounding boxes of the objects. Then, with these depth maps and  bounding boxes of objects, they obtain the depths of objects. Since the accuracy of depths of objects heavily depends on the efficiency of two pre-trained models, the computed depth information is unreliable. Besides, the introducing of  3D information increases the complexity of model. Even so, leveraging 3D knowledge is another idea to overcome the influence of viewpoint changes. This inspires us to further explore this task in the future.  

 However, the aforementioned methods capture changed features between two images mainly based on the global (direct subtraction) or local (individual feature matching) level, while not trying to learn the features of change based on the neighborhood level. In addition, the problem of syntax ambiguous in ground-truth captions are disregarded.  Instead, we propose a neighborhood contrastive transformer. It compares two images at neighborhood level to first capture differentiating properties from each image and then learn contrastive information between them. In addition, it employs dependency relations to solve the problem of structure ambiguity in change captions.


\textbf{Contrastive Feature learning in Captioning.}
Learning contrastive features is to model similar/dissimilar image
representations from similar/dissimilar image pairs \cite{liu2021contrastive}. This idea has been attempted by recent works in group captioning \cite{li2020context} and  chest X-ray report generation \cite{liu2021contrastive}. On the one hand, given two groups of images, Li \emph{et al.} \cite{li2020context} propose to use self-attention mechanism to capture common properties from each image group and then capture contrastive information between them. On the other hand,  given a chest X-ray image and a set of norm images, Liu \emph{et al.} \cite{liu2021contrastive} present a contrastive attention model to learn contrastive features between the input image with normal images.
There are two major differences between our method and them. First, there exist irrelevant distractors (\emph{e.g.,} viewpoint change) in our task, which brings the additional challenge to distinguish real change from pseudo change. Second, different from them matching feature individually, our method is first to aggregate neighboring features, and then perform feature matching at neighborhood level to construct contrastive features, which aims to identify fine-grained change while being immune to viewpoint change.  

\textbf{Syntax Knowledge Used in Captioning.} There have been some attempts that use the syntax knowledges of Part-of-Speech (PoS) and syntax dependencies between words in captioning. 
On the one hand, Hou \emph{et al.} \cite{Hou_2019_ICCV} propose to model the syntactic structure and exploit the semantic primitive by learning the joint probability of the PoS sequence and words. Wang \emph{et al.} \cite{wang2019controllable} present a PoS generator to predict the global syntactic PoS information of sentences. Zhang \emph{et al.} \cite{Zhang2021IntegratingPO} and Deng \emph{et al.} \cite{9367203} propose to make the model adaptively generate each word based on its PoS, thus improving the cross-modal alignment. On the other hand,  Zheng \emph{et al.} \cite{zheng2020syntax} propose to decode syntax components (subject, object and predicate) for targeting the action in video clips. Zhao \emph{et al.} \cite{zhao2021multi} devise a multi-modal dependency tree construction approach to capture the syntactic and semantic dependencies in long and complex video captions. 

In change captioning, most works focus on learning an accurate change representation for caption generation, while ignoring the exploitation of syntax knowledge. Similar to Zhang \emph{et al.} and Deng \emph{et al.}, Tu \emph{et al.} \cite{tu2021semantic} introduce PoS information and propose an attention-based visual switch to dynamically use visual information. Different from this work, we aim to exploit explicit syntax dependencies between words to disambiguate syntax structure of change captions, which is beneficial to help the model differentiate changed object and its referent in ground-truth captions during training. 

\begin{figure*}[t]
\centering
\includegraphics[width=1\textwidth]{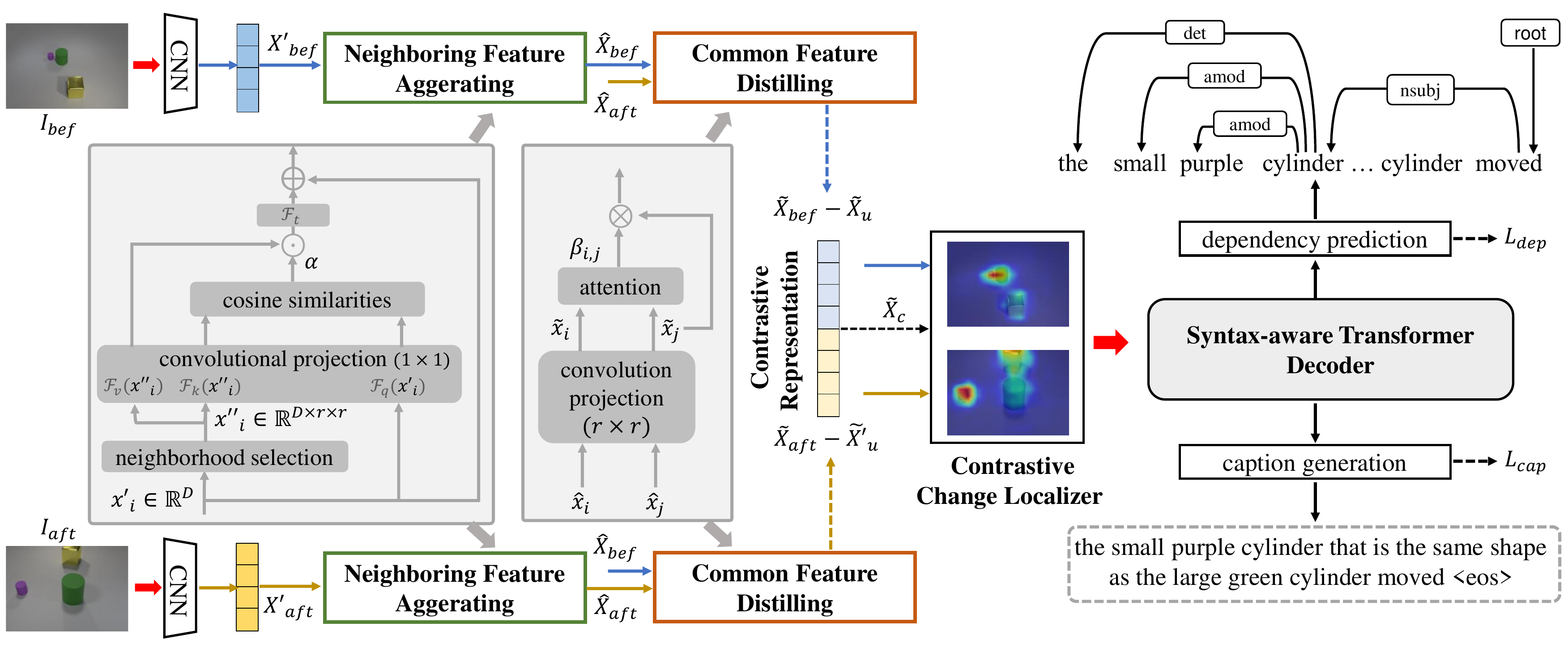} 
  \caption{ The architecture of the proposed neighborhood contrastive transformer, including a neighborhood feature aggregating, a common feature distilling, a contrastive change localizer and a syntax-aware transformer decoder.}
\label{fig2}
\end{figure*}

\section{Methodology}
\label{method}
As shown in Fig. \ref{fig2}, the architecture of our method consists of four parts: (1) a neighborhood feature aggregating module identifies the fine-grained change and resists irrelevant viewpoint changes; (2) a common feature distilling module extracts differentiating information from each image, and learns contrastive information between them; (3) a contrastive change localizer locates the specific change features on the two images; (4) a syntax-aware transformer decoder translates the learned features of change into a natural language sentence, and predicts the syntax dependencies between words. 


\subsection{Neighborhood Feature Aggregating}
Formally, given two images of ``before'' $I_{bef}$ and ``after'' $I_{aft}$, we exploit an off-the-shelf CNN to extract grid features for them, denoted as $X_{bef}$ and $X_{aft}$, where $X \in \mathbb{R}^{C \times H \times W}$. C, H, W indicate the number of channels, height, and width. 
Although the CNN  can capture local spatial context, these correlations are modeled based on single image without viewpoint changes and cannot be directly transferred to change captioning. Besides, the latest work \cite{zhao2021multi2} in semantic correspondence shows that local self-attention performs  well in capturing relations between neighboring elements. Inspired by this, we design a neighborhood feature aggregating module  to dynamically update each feature by integrating spatial context from the same neighborhood of two images.

Concretely, for ${X_{bef(aft)}}=\left\{{x}_{1}, \ldots, {x}_{N}\right\}$ ($N=HW$), where $x_{i}\in\mathbb{R}^{C}$, we first project the feature of $i$-th grid cell $x_{i}$ into a low-dimensional embedding space of $\mathbb{R}^{D}$ by a shared linear transformation:
\begin{equation}
{x}_{i}^{\prime}={M}_{v} {x}_{i}+ {b}_{v} + pos(\tilde x,\tilde y),
\end{equation}
where ${M}_{v} \in \mathbb{R}^{D \times C}$ and ${b}_{v} \in \mathbb{R}^{D}$ are trainable parameters.  pos($\tilde x,\tilde y) \in \mathbb{R}^{D}$ is a learnable position embedding for $i$-th grid feature. Herein, $\tilde{x}$ and $\tilde{y}$ are the orders of each feature in the height and width dimensions of an image. Position embedding layers are two lookup tables of size (H, D/2) and size (W, D/2). Based on the orders of each feature, we can learn its position embeddings from height and width dimensions, and concatenate them as its position embedding. 
Then, for every grid feature $x'_{i}$, we pick out its $r \times r$ neighboring features and acquire a neighborhood feature representation $X''_{bef(aft)}\in \mathbb{R}^{C \times H \times W \times r \times r}$. Next, we measure feature cosine similarities between $x'_{i}$ and its  $r \times r$ neighborhood:
\begin{equation}
\begin{aligned}
\label{weight}
e =\operatorname{\Phi} \left [\mathcal{F}_{q}\left({x}'_{i}\right), \mathcal{F}_{k}\left({x}''_{i}\right)\right],\\
\alpha \sim \operatorname{Softmax}\left(e\right),
\end{aligned}
\end{equation}
where $\alpha \in \mathrm{R}^{r \times r}$ is the relation coefficient indicating how much message to obtain from the neighboring features; $\Phi$ is the cosine similarity function. $\mathcal{F}_{q} \in \mathbb{R}^D$ and  $\mathcal{F}_{k} \in \mathbb{R}^{D \times r^{2}}$ are two convolution layers. 
Finally, $x'_i$ is updated to $\hat x_i$ via aggregating related information from the neighboring features:
\begin{equation}
\hat x_i={x}'_{i} + \mathcal{F}_{t}(\sum_{r, r} \alpha \odot \mathcal{F}_{v}\left({x}''_{i}\right)), \hat x_i \in \mathbb{R}^{D},
\end{equation}
where $\odot$ refers to element-wise multiplication.
The above operation updates the original features into 
$\hat X_{bef}$ and $\hat X_{aft}$, which enables the model to identify inconspicuous and occluded changes, while being immune to viewpoint change.

\subsection{Common Feature Distilling}
As shown in Fig. \ref{fig1}, compared to the tiny change, most properties are identical between the image pair. Hence, it is natural to find and remove the common portion from the two images, and the remaining information can be treated as contrastive features. Motivated by this, a common feature distilling module is designed to compare two images and learn an effective contrasive representation.

Herein, we learn the change features in $\hat X_{bef}$ compared to  $\hat X_{aft}$. In detail, we first exploit a shared transformation layer with depth-wise separate convolutions to project $\hat X_{bef}$ and $\hat X_{aft}$ into a common semantic space. This further captures spatial correlations in the same neighborhood:
\begin{equation}
\tilde X_{bef(aft)}=\mathcal{F}_{depth}\left(\hat X_{bef(aft)}, s\right), \tilde X \in \mathbb{R}^{D \times H \times W},
\end{equation}
where $s$ is with the kernel size of  $r \times r$, and we reshape $\tilde X_{bef(aft)}$ to $\tilde X_{bef(aft)}\in\mathbb{R}^{N \times D}$. Then, we measure the  similarity between every feature $\tilde {x}_{i}^{b} $ in $\tilde X_{bef}$  and every  feature $\tilde {x}_{j}^{a} $ in $\tilde X_{aft}$ by the dot-product attention:
\begin{equation}
\beta_{i,j}=\frac{\exp \left(\beta_{i,j}^{\prime}\right)}{\sum_{ j} \exp \left(\beta_{i,j}^{\prime}\right)}, \quad \beta_{i,j}^{\prime}=\tilde {x}_{i}^{bT} \tilde {x}_{j}^{a},
\end{equation}
where $\beta_{i,j} \in {B}$ is a set of similarity scores to indicate which features are the common properties between  $\tilde X_{bef}$ and $\tilde X_{aft}$.
Then, the common features are extracted from $\tilde X_{aft}$ under the guidance of the learned similarity score matrix $B$ : 
\begin{equation}
\tilde X_{u}= B \cdot \tilde X_{aft}.
\end{equation}
Next, we remove the common features $\tilde X_{u}$ from $\tilde X_{bef}$ to distill the change features:
\begin{equation}
  \tilde X_{c}^{bef} =  \tilde X_{bef} - \tilde X_{u}.
\end{equation}
By that analogy, we distill the change features $\tilde X_{c}^{aft}$ in $\tilde X_{aft}$ with reference to $\tilde X_{bef}$. Finally, we construct the contrastive representation between two images by fusing the above change features, which is implemented by a fully-connected layer with the ReLU activation function:
\begin{equation}
\tilde{X}_{c}=\operatorname{ReLU}\left(\left[\tilde{X}_{c}^{b e f} ; \tilde{X}_{c}^{a f t}\right] W_{h} + b_h \right ).
\end{equation}

\subsection{Contrastive Change Localizer}
\label{localizer}
After learning the contrastive representation $\tilde{X}_{c}$, we introduce a contrastive change localizer based on spatial attention mechanism, which is used to pinpoint change on the two images. 
Concretely, it first generates two attention maps by using $\tilde{X}_{c}$ to query each image representation, respectively:
\begin{equation}
\begin{array}{c}
\gamma_{bef} =\sigma\left(M L P\left(\left[\tilde X_{c}; \tilde X_{bef} \right]\right)\right), \\
\gamma_{aft} =\sigma\left(M L P\left(\left[\tilde X_{c}; \tilde X_{aft} \right]\right)\right),
\end{array}
\end{equation}
where $MLP$ is a two-layer multi-layer perceptron with the ReLU activation function in between. [;] and $\sigma$ denote concatenation operation and sigmoid activation function.
Further, the specific feature of change is localized via implementing a weighted-sum pooling on each image representation over the spatial dimensions, respectively:
\begin{equation}
\begin{array}{c}
l_{{bef}}=\sum_{H, W} \gamma_{\mathrm{bef}} \odot \tilde X_{{bef}}, l_{{bef}} \in \mathbb{R}^{D}, \\
l_{{aft}}=\sum_{H, W} \gamma_{\mathrm{aft}} \odot \tilde X_{{aft}}, l_{\mathrm{aft}} \in \mathbb{R}^{D}.
\end{array}
\end{equation}

\subsection{Syntax-aware Language Decoder}

With the pooling change features $l_{b e f},l_{a f t}$, and their difference feature ${l_{d i f f}}$, we first concatenate them as $ V \in \mathbb{R}^{3 \times D}$. Then, the decoder of transformer learns the cross-modal alignment between the word embedding features $ E[W]=\{E[w_1],... ,E[w_{m}]\}$ and visual features $ V $.  Finally, the decoder exploits attended features of change to generate sentences, during which we introduce the syntax knowledge of dependencies between words to calibrate the decoder. This aims to solve the problem of syntax ambiguity in change descriptions.

\subsubsection{Background Knowledge}
We first briefly review the framework of standard transformer decoder. The key module is the scaled dot-product attention.  Given a query matrix $Q \in \mathbb{R}^{T_{q} \times d_{k}},$ key matrix $K \in \mathbb{R}^{T_{v} \times d_{k}}$ and value matrix $V \in \mathbb{R}^{T_{v} \times d_{v}},$ the
attention result is computed as:
 \begin{equation}
\text {Attention }(Q, K, V)=\operatorname{Softmax}\left(\frac{Q K^{\top}}{\sqrt{d_{k}}}, \operatorname{dim}=1\right) V.
\end{equation}
The multi-head attention is based on the scaled dot-product attention. It consists of $h$ different ``heads". For each head, the attention result is computed by:
 \begin{equation}
\text { head }_{i}=\operatorname{Attention}\left(Q W_{i}^{Q}, K W_{i}^{K}, V W_{i}^{V}\right).
\end{equation}
Afterward, the multi-head attention operation is to concatenate all the heads, which is defined as:
 \begin{equation}
\text { MultiHead }(Q, K, V)=\text { Concat }_{i=1 \ldots h}\left(\text { head }_{i}\right) W^{O}.
\end{equation}
Further, the output of each attention layer $x$ is fed into a feed-forward network (FFN) based on a non-linear transformation:
\begin{equation}
\operatorname{FFN}(\boldsymbol{x})=\operatorname{GELU}\left(\boldsymbol{x} W_{f1}+\boldsymbol{b}_{f1}\right) W_{f2}+\boldsymbol{b}_{f2}.
\end{equation}

Then, we give a introduction about the dependencies between words.  In natural language processing, dependency parsing refers to the process of examining the dependencies between the linguistic units (\emph{e.g.}, words) of a sentence, in order to determine its grammatical structure. That is, syntax dependency is the notion that words are connected to each other by directed links. The verb is taken to be the structural center of clause structure and tagged as ``root''. All other syntactic words are either directly or indirectly connected to the ``root'' in terms of the directed links.  In Fig. \ref{fig2}, we mainly illustrate the main words of this change caption. A dependence tag indicates the relationship between two words. For example, the word ``moved'' changes the meaning of the noun ``cylinder''. Therefore, we can find that a dependency from ``moved'' to ``cylinder'', where ``moved'' is the pinnacle and ``cylinder'' is the kid or dependent. The tag of this dependency is ``nsubj'', which stands for nominal subject of this sentence. The verb ``moved'' is the root in this dependency structure. In addition, we notice that there is no directed link between the other ``cylinder'' and the ``moved''. Based on these directed links, the model can better understand complex structure in a sentence and thus identify which object changed.  

\subsubsection{Decoding Stage}
The decoder contains a stack of $N$ identical layers. At the $l$-th decoder layer,  the masked self-attention layer, which prevents the model from seeing future words, first takes the word embedding features $ E[W]=\{E[w_1],... ,E[w_{m}]\}$ as the inputs and models their relationships. The operation is defined as:
\begin{equation}
 {\hat E[W]}=\text { LN }(E[W]+\text { MultiHead }(E[W],E[W],E[W])),
\end{equation}
where LN is short for layer normalization \cite{ba2016layer}.
Then, the decoder utilizes the attended features $ \hat E[W]$ to query the most related features from $ V $ based on the cross-attention layer:
\begin{equation}
\hat H=\text { LN } ({E[\hat W]}+\text { MultiHead }({E[\hat W]},V, V)).
\end{equation}
Afterward, the $\hat H$ is passed to a feed-forward layer:
\begin{equation}
\tilde H=\text { LN}(\hat H+\operatorname {FFN}(\hat H)).
\end{equation}
Finally, the probability distributions of target words and dependencies are calculated via two separate single hidden layers:
\begin{equation}
\begin{aligned}
W=\operatorname{Softmax}\left(\tilde H W_{c}+{b}_{c}\right), \\ 
D =\operatorname{Softmax}\left(\tilde H W_{d}+{b}_{d}\right),
\end{aligned}
\end{equation}
where $W_{c}\in \mathbb{R}^{D \times U}$, $W_{d}\in \mathbb{R}^{D \times n}$, $b_{c} \in \mathbb{R}^{U}$, and $b_{d} \in \mathbb{R}^{n}$ are the parameters to be learned. $U$ is the dimension of vocabulary size; $n$ is the number of dependency relations.

\subsection{Joint Training}
 We jointly train the caption generator and dependency predictor in an end-to-end manner by maximizing the likelihood of the observed word sequences and dependency relations. Given the target ground-truth caption words $\left(w_{1}^{c}, \ldots, w_{m}^{c}\right)$ and dependency relations $\left(w_{1}^{d}, \ldots, w_{m}^{d}\right)$, we minimize the negative log-likelihood loss of caption generator and dependency predictor, respectively:
 
\begin{equation}
\begin{aligned}
L_{cap}(\theta_c)=-\sum_{t=1}^{m} \log p\left(w_{t}^{c} \mid w_{<t}^{c}; \theta_c\right),\\
L_{dep}(\theta_d)=-\sum_{t=1}^{m} \log p\left(w_{t}^{d} \mid w_{<t}^{d}; \theta_d\right),
\end{aligned}
\end{equation}
where $\theta_c$ and $\theta_p$ are the parameters of the caption generator and dependency predictor, respectively. $m$ is the length of the caption and dependencies.
The final loss function is optimized as follows:
\begin{equation}
\label{cross-entropy}
L(\theta)=L_{cap} + \lambda L_{dep},
\end{equation}
where $\lambda$ is a trade-off parameter to balance the contributions from the caption generator and dependency predictor.

\section{Experiments}
\label{experiment}
\subsection{Datasets}
\textbf{Image Editing Request} dataset \cite{tan2019expressing} is comprised of 3,939 real image pairs with 5,695 editing  instructions.  Each image pair in the training set
has one instruction, and each image pair in the validation
and test sets has three instructions. The changed objects in this dataset are usually inconspicuous and vague. We use the official split with 3,061 image pairs for training, 383 for validation, and 495 for testing.

\textbf{CLEVR-Change} is a large-scale synthetic dataset \cite{park2019robust} with moderate viewpoint change. It has 79,606 image pairs and 493,735 captions, including five change types, \emph{i.e.}, ``Color'', ``Texture'', ``Add'', ``Drop'', and ''Move''. It has two change settings: both scene and pseudo change and only pseudo change. We use the official split with 67,660 for training, 3,976 for validation and 7,970 for testing.

\textbf{CLEVR-DC} is a large-scale synthetic dataset \cite{kim2021agnostic} to simulate extreme viewpoint shifts. It consists of 48,000 pairs with the same change types as CLEVR-Change. We use the official split with  85\% for training, 5\% for validation, and 10\% for test, respectively. 

\subsection{Evaluation Metrics}
Following the state-of-the-art methods \cite{shi2020finding,tu-etal-2021-r,hosseinzadeh2021image}, five metrics are used to evaluate the generated sentences, \emph{i.e.}, BLEU-4 (B) \cite{papineni2002bleu}, METEOR (M) \cite{banerjee2005meteor}, ROUGE-L (R) \cite{lin2004rouge}, CIDEr (C) \cite{vedantam2015cider}, and SPICE (S) \cite{anderson2016spice}. BLEU-4 is exploited for corpus level comparisons of 4-gram matches and has been widely used in machine translation task. METEOR is designed to measure the relationship between candidate and reference sentences
based on exact token matching. ROUGE-L computes the word correlations  that co-exist in two sentences in the same order, based on the Longest Common Sub-sequence (LCS). CIDEr is recently proposed and especially designed for the captioning task to capture human judgment of consensus. SPICE is also a new metric and designed for captioning task, which compares semantic propositional content between candidate and reference sentences. We compute results based on the Microsoft COCO evaluation server \cite{chen2015microsoft}.

\subsection{Implementation Details}
For a fair comparison, we follow the state-of-the-art methods to use a ResNet-101 model \cite{he2016deep} pre-trained on the Imagenet dataset \cite{russakovsky2015imagenet} for extracting grid features of an image pair, with the dimension of 1024 $\times$ 14 $\times$ 14. We first project these  features into a lower dimension of 512. The hidden size in the overall model and word embedding size in the decoder are set to 512 and 300.  To obtain the ground-truth dependencies, we exploit a pre-trained Biaffine Parse \cite{dozat2016deep}  to extract the explicit dependency relations of each sentence in the training sets.   We set the layer number of neighborhood feature aggregating as 1; the number of dependency tags as 49; the neighborhood range $r$ as 3; the layer number of decoder as 2; the number of attention head as 8. 

During training, on CLEVR-Change and CLEVR-DC, we set the batch size and learning rate as 128 and 2 $\times$ $10^{-4}$. On Image Editing Request, the batch size and learning rate are set to 32 and 2 $\times$ $10^{-4}$. We use Adam optimizer \cite{kingma2014adam} to minimize the negative log-likelihood loss of Eq. (\ref{cross-entropy}). In the inference phase, the greedy decoding strategy is used to generate target captions. Both training and inference are implemented with PyTorch \cite{paszke2019pytorch} on an RTX 3090 GPU.

\begin{table}[t]
  \centering
  \caption{Comparison with the state-of-the-art methods on CLEVR-Change on Total Performance. ``*'' represents this model is trained with three pre-training tasks.}
    \begin{tabular}{c|c|c|c|c|c}
    \hline
          & \multicolumn{5}{c}{Total}  \\
    \hline
    Method & B & M & R & C	 & S  \\
    \hline
    DUDA (ICCV'19) & 47.3  & 33.9  & -     & 112.3 & 24.5 \\
    M-VAM (ECCV'20) & \cellcolor[rgb]{ .953,  .976,  .929}50.3 & \cellcolor[rgb]{ .953,  .976,  .929}37.0 & \cellcolor[rgb]{ .953,  .976,  .929}69.7 & 114.9 & \cellcolor[rgb]{ .953,  .976,  .929}30.5 \\
    DUDA+TIRG (CVPR'21) & \cellcolor[rgb]{ .886,  .937,  .855}51.2 & \cellcolor[rgb]{ .953,  .976,  .929}37.7 & \cellcolor[rgb]{ .886,  .937,  .855}70.5 & \cellcolor[rgb]{ .953,  .976,  .929}115.4 & \cellcolor[rgb]{ .886,  .937,  .855}31.1 \\
    IFDC (TMM'21) & \cellcolor[rgb]{ .953,  .976,  .929}49.2 & 32.5  & 69.1  & \cellcolor[rgb]{ .953,  .976,  .929}118.7 & - \\
    R$^{3}$Net+SSP (EMNLP'21) & \cellcolor[rgb]{ .776,  .878,  .706}54.7 & \cellcolor[rgb]{ .886,  .937,  .855}39.8 & \cellcolor[rgb]{ .776,  .878,  .706}73.1 & \cellcolor[rgb]{ .776,  .878,  .706}123.0 & \cellcolor[rgb]{ .776,  .878,  .706}\underline{32.6} \\
    VACC (ICCV'21) & \cellcolor[rgb]{ .886,  .937,  .855}52.4 & \cellcolor[rgb]{ .953,  .976,  .929}37.5 & -     & 114.2 & \cellcolor[rgb]{ .886,  .937,  .855}31.0 \\
    SRDRL+AVS (ACL'21) & \cellcolor[rgb]{ .776,  .878,  .706}54.9 & \cellcolor[rgb]{ .776,  .878,  .706}\underline{40.2} & \cellcolor[rgb]{ .776,  .878,  .706}73.3 & \cellcolor[rgb]{ .886,  .937,  .855}122.2 & \cellcolor[rgb]{ .776,  .878,  .706}\textbf{32.9} \\
    SGCC (ACM MM'21) & \cellcolor[rgb]{ .776,  .878,  .706}51.1 & \cellcolor[rgb]{ .663,  .816,  .557}\textbf{40.6} & \cellcolor[rgb]{ .663,  .816,  .557}\textbf{73.9} & \cellcolor[rgb]{ .886,  .937,  .855}121.8 & \cellcolor[rgb]{ .776,  .878,  .706}32.2 \\
    MCCFormers-D (ICCV'21) & \cellcolor[rgb]{ .886,  .937,  .855}52.4 & \cellcolor[rgb]{ .886,  .937,  .855}38.3 & -     & \cellcolor[rgb]{ .886,  .937,  .855}121.6 & 26.8 \\
    PCL w/o PT (AAAI'22) & 32.7  & 27.7  & 57.2  & 89.8  & - \\
    PCL w/ PT (AAAI'22) * & \cellcolor[rgb]{ .886,  .937,  .855}51.2 & \cellcolor[rgb]{ .953,  .976,  .929}36.2 & \cellcolor[rgb]{ .886,  .937,  .855}71.7 & \cellcolor[rgb]{ .663,  .816,  .557}\textcolor[rgb]{ .322,  .322,  .322}{\textbf{128.9}} & - \\
    \hline
    NCT (Ours) & \cellcolor[rgb]{ .663,  .816,  .557}\textbf{55.1} & \cellcolor[rgb]{ .776,  .878,  .706}\underline{40.2} & \cellcolor[rgb]{ .663,  .816,  .557}\underline{73.8} & \cellcolor[rgb]{ .663,  .816,  .557}\textbf{124.1} & \cellcolor[rgb]{ .663,  .816,  .557}\textbf{32.9} \\
    \hline
    \end{tabular}%
  \label{clevr-total_com}%
\end{table}%

\begin{table}[t]
  \centering
  \caption{Comparison with the state-of-the-art methods on CLEVR-Change on scene change.}
    \begin{tabular}{c|c|c|c|c|c}
    \hline
          & \multicolumn{5}{c}{Scene Change}  \\
    \hline
    Method & B     & M     & R     & C     & S  \\
    \hline
    DUDA (ICCV'19) & 42.9  & 29.7  & -     & 94.6  & 19.9  \\
    DUDA+TIRG (CVPR'21) & \cellcolor[rgb]{ .878,  .953,  .984}49.9 & \cellcolor[rgb]{ .878,  .953,  .984}34.3 & \cellcolor[rgb]{ .878,  .953,  .984}65.4 & \cellcolor[rgb]{ .878,  .953,  .984}101.3 & \cellcolor[rgb]{ .878,  .953,  .984}27.9 \\
    IFDC (TMM'21) & \cellcolor[rgb]{ .878,  .953,  .984}47.2 & \cellcolor[rgb]{ .957,  .976,  .984}29.3 & 63.7  & \cellcolor[rgb]{ .878,  .953,  .984}105.4 & - \\
    R$^{3}$Net+SSP (EMNLP'21)  & \cellcolor[rgb]{ .867,  .922,  .969}52.7 & \cellcolor[rgb]{ .867,  .922,  .969}36.2 & \cellcolor[rgb]{ .867,  .922,  .969}69.8 & \cellcolor[rgb]{ .741,  .843,  .933}116.6 & \cellcolor[rgb]{ .867,  .922,  .969}30.3 \\
    SRDRL+AVS (ACL'21) & \cellcolor[rgb]{ .867,  .922,  .969}52.7 & \cellcolor[rgb]{ .867,  .922,  .969}36.4 & \cellcolor[rgb]{ .867,  .922,  .969}69.7 & \cellcolor[rgb]{ .867,  .922,  .969}114.2 & \cellcolor[rgb]{ .867,  .922,  .969}30.8  \\
    \hline
    NCT (Ours)  & \cellcolor[rgb]{ .608,  .761,  .902}\textbf{53.1} & \cellcolor[rgb]{ .608,  .761,  .902}\textbf{36.5} & \cellcolor[rgb]{ .608,  .761,  .902}\textbf{70.7} & \cellcolor[rgb]{ .608,  .761,  .902}\textbf{118.4} & \cellcolor[rgb]{ .741,  .843,  .933}\textbf{30.9}  \\
    \hline
    \end{tabular}%
  \label{clevr-sc-com}%
\end{table}%

\begin{table}[t]
  \centering
  \caption{A detailed breakdown of evaluation on CIDEr with different change types: ``(C) Color'', ``(T) Textur'', ``(A) Add'', ``(D) Drop'',
and ``(M) Move''.}
    \begin{tabular}{c|c|c|c|c|c}
    \hline
          & \multicolumn{5}{c}{CIDEr}  \\
    \hline
    Method & C & T & A   & D  & M  \\
    \hline
    DUDA (ICCV'19) & 120.4 & 86.7  & 108.3 & 103.4 & 56.4  \\
    M-VAM (ECCV'20) & \cellcolor[rgb]{ 1,  .933,  .859}122.1 & \cellcolor[rgb]{ 1,  .933,  .859}98.7 & \cellcolor[rgb]{ .988,  .894,  .839}126.3 & \cellcolor[rgb]{ 1,  .933,  .859}115.8 & \cellcolor[rgb]{ .988,  .894,  .839}82.0 \\
    DUDA+TIRG (CVPR'21) & 120.8 & 89.9  & \cellcolor[rgb]{ 1,  .933,  .859}119.8 & \cellcolor[rgb]{ .973,  .796,  .678}123.4 & 62.1 \\
    IFDC (TMM'21) & \cellcolor[rgb]{ .988,  .894,  .839}133.2 & \cellcolor[rgb]{ 1,  .933,  .859}99.1 & \cellcolor[rgb]{ .973,  .796,  .678}128.2 & \cellcolor[rgb]{ .988,  .894,  .839}118.5 & \cellcolor[rgb]{ .988,  .894,  .839}82.1 \\
    R$^{3}$Net+SSP (EMNLP'21) & \cellcolor[rgb]{ .973,  .796,  .678}139.2 & \cellcolor[rgb]{ .973,  .796,  .678}123.5 & \cellcolor[rgb]{ 1,  .933,  .859}122.7 & \cellcolor[rgb]{ .988,  .894,  .839}121.9 & \cellcolor[rgb]{ .957,  .69,  .518}\textbf{88.1} \\
    SRDRL+AVS (ACL'21) & \cellcolor[rgb]{ .988,  .894,  .839}136.1 & \cellcolor[rgb]{ .973,  .796,  .678}122.7 & \cellcolor[rgb]{ 1,  .933,  .859}121.0 & \cellcolor[rgb]{ .973,  .796,  .678}126.0 & \cellcolor[rgb]{ 1,  .933,  .859}78.9 \\
    SGCC (ACM MM'21) & \cellcolor[rgb]{ 1,  .933,  .859}128.0 & \cellcolor[rgb]{ .973,  .796,  .678}122.9 & \cellcolor[rgb]{ 1,  .933,  .859}117.1 & \cellcolor[rgb]{ 1,  .933,  .859}116.9 & \cellcolor[rgb]{ 1,  .933,  .859}77.1 \\
    PCT w/ PT (AAAI'22) & \cellcolor[rgb]{ .988,  .894,  .839}131.2 & \cellcolor[rgb]{ 1,  .933,  .859}101.1 & \cellcolor[rgb]{ .957,  .69,  .518}\textbf{133.3} & \cellcolor[rgb]{ 1,  .933,  .859}116.5 & \cellcolor[rgb]{ 1,  .933,  .859}81.7  \\ \hline
    NCT (Ours)  & \cellcolor[rgb]{ .957,  .69,  .518}\textbf{140.2} & \cellcolor[rgb]{ .957,  .69,  .518}\textbf{128.8} & \cellcolor[rgb]{ .973,  .796,  .678} \underline{128.4} & \cellcolor[rgb]{ .957,  .69,  .518}\textbf{129.0} & \cellcolor[rgb]{ .973,  .796,  .678}\underline{86.0}  \\
    \hline
    \end{tabular}%
  \label{type_cider}%
\end{table}%

\begin{table}[t]
  \centering
  \caption{A detailed breakdown of evaluation on SPICE.}
    \begin{tabular}{c|c|c|c|c|c}
    \hline
          & \multicolumn{5}{c}{SPICE}  \\
    \hline
    Method & C & T & A   & D  & M  \\
    \hline
    DUDA (ICCV'19) & 21.2  & 18.3  & 22.4  & 22.2  & 15.4  \\
    M-VAM (ECCV'20) & \cellcolor[rgb]{ 1,  .933,  .859}28.0 & \cellcolor[rgb]{ 1,  .933,  .859}26.7 & \cellcolor[rgb]{ 1,  .933,  .859}30.8 & \cellcolor[rgb]{ .973,  .796,  .678}32.3 & \multicolumn{1}{c}{\cellcolor[rgb]{ 1,  .933,  .859}22.5} \\
    DUDA+TIRG (CVPR'21) & \cellcolor[rgb]{ .988,  .894,  .839}29.7 & \cellcolor[rgb]{ .988,  .894,  .839}27.4 & \cellcolor[rgb]{ 1,  .933,  .859}31.4 & \cellcolor[rgb]{ 1,  .933,  .859}30.8 & \cellcolor[rgb]{ .988,  .894,  .839}23.5 \\
    R$^{3}$Net+SSP (EMNLP'21) & \cellcolor[rgb]{ .973,  .796,  .678}31.6 & \cellcolor[rgb]{ .973,  .796,  .678}30.8 & \cellcolor[rgb]{ .973,  .796,  .678}\underline{32.3} & \cellcolor[rgb]{ .988,  .894,  .839}31.7 & \cellcolor[rgb]{ .973,  .796,  .678}25.4 \\
    SRDRL+AVS (ACL'21) & \cellcolor[rgb]{ .957,  .69,  .518}\textbf{32.4} & \cellcolor[rgb]{ .973,  .796,  .678}30.9 & \cellcolor[rgb]{ .957,  .69,  .518}\textbf{33.0} & \cellcolor[rgb]{ .973,  .796,  .678}32.4 & \cellcolor[rgb]{ .973,  .796,  .678}25.4  \\ 
   SGCC (ACM MM'21) & \cellcolor[rgb]{ .988,  .894,  .839}30.0 & \cellcolor[rgb]{ .973,  .796,  .678}31.1 & \cellcolor[rgb]{ 1,  .933,  .859}30.8 & \cellcolor[rgb]{ 1,  .933,  .859}30.1 & \cellcolor[rgb]{ .973,  .796,  .678}25.3 \\
    
    \hline
    NCT (Ours)  & \cellcolor[rgb]{ .957,  .69,  .518}\textbf{32.4} & \cellcolor[rgb]{ .957,  .69,  .518}\textbf{31.8} & \cellcolor[rgb]{ .973,  .796,  .678}\underline{32.3} & \cellcolor[rgb]{ .957,  .69,  .518}\textbf{32.6} & \cellcolor[rgb]{ .957,  .69,  .518}\textbf{25.5}  \\
    \hline
    \end{tabular}%
  \label{type_spice}%
\end{table}%

\begin{table}[t]
  \centering
  \caption{A detailed breakdown of evaluation on METEOR.}
    \begin{tabular}{c|c|c|c|c|c}
    \hline
          & \multicolumn{5}{c}{METEOR}  \\
    \hline
    Method & C & T & A   & D  & M  \\
    \hline
    DUDA (ICCV'19) & 32.8  & 27.3  & 33.4  & \cellcolor[rgb]{ 1,  .933,  .859}31.4 & 23.5 \\
    M-VAM+RAF (ECCV'20) & \cellcolor[rgb]{ 1,  .933,  .859}35.8 & \cellcolor[rgb]{ .988,  .894,  .839}32.3 & \cellcolor[rgb]{ .988,  .894,  .839}37.8 & \cellcolor[rgb]{ .988,  .894,  .839}36.2 & \cellcolor[rgb]{ 1,  .933,  .859}27.9 \\
    DUDA+TIRG (CVPR'21) & \cellcolor[rgb]{ .988,  .894,  .839}36.1 & \cellcolor[rgb]{ 1,  .933,  .859}30.4 & \cellcolor[rgb]{ .988,  .894,  .839}37.8 & \cellcolor[rgb]{ .973,  .796,  .678}36.7 & \cellcolor[rgb]{ 1,  .933,  .859}27.0 \\
    IFDC (TMM'21) & \cellcolor[rgb]{ 1,  .933,  .859}33.1 & \cellcolor[rgb]{ 1,  .933,  .859}27.9 & \cellcolor[rgb]{ 1,  .933,  .859}36.2 & \cellcolor[rgb]{ 1,  .933,  .859}31.4 & \cellcolor[rgb]{ .973,  .796,  .678}31.2 \\
    R$^{3}$Net+SSP (EMNLP'21)  & \cellcolor[rgb]{ .957,  .69,  .518}38.9 & \cellcolor[rgb]{ .973,  .796,  .678}35.5 & \cellcolor[rgb]{ .973,  .796,  .678}38.0 & \cellcolor[rgb]{ .973,  .796,  .678}37.5 & \cellcolor[rgb]{ .973,  .796,  .678}30.9 \\
    SRDRL+AVS (ACL'21) & \cellcolor[rgb]{ .957,  .69,  .518}39.0 & \cellcolor[rgb]{ .973,  .796,  .678}35.6 & \cellcolor[rgb]{ .973,  .796,  .678}38.9 & \cellcolor[rgb]{ .957,  .69,  .518}\textbf{38.0} & \cellcolor[rgb]{ .988,  .894,  .839}30.1 \\
    SGCC (ACM MM'21) & \cellcolor[rgb]{ .988,  .894,  .839}37.8 & \cellcolor[rgb]{ .973,  .796,  .678}36.1 & \cellcolor[rgb]{ .973,  .796,  .678}38.9 & \cellcolor[rgb]{ 1,  .933,  .859}36.7 & \cellcolor[rgb]{ .957,  .69,  .518}\textbf{32.8} \\ \hline
 
    NCT (Ours)  & \cellcolor[rgb]{ .957,  .69,  .518}\textbf{39.1} & \cellcolor[rgb]{ .957,  .69,  .518}\textbf{36.3} & \cellcolor[rgb]{ .957,  .69,  .518}\textbf{39.0} & \cellcolor[rgb]{ .973,  .796,  .678}37.2 & \cellcolor[rgb]{ .973,  .796,  .678}30.5 \\ \hline
    
    \end{tabular}%
  \label{type_meteor}%
\end{table}%

\subsection{Performance Comparison}
\subsubsection{Results on the CLEVR-Change Dataset.}
We compare the proposed method with the state-of-the-art methods in: 1) total performance evaluating the overall performance under both scene and pseudo changes; 2) scene change; 3) different change types. The ten comparison methods are DUDA \cite{park2019robust}, M-VAM \cite{shi2020finding}, IFDC \cite{huang2021image}, DUDA+TIRG \cite{hosseinzadeh2021image}, R$^{3}$Net+SSP \cite{tu-etal-2021-r}, VACC \cite{kim2021agnostic}, SRDRL+AVS \cite{tu2021semantic}, MCCFormers-D \cite{Qiu_2021_ICCV}, SGCC \cite{liao2021scene}, and PCL w/ and w/o PT (pre-training) \cite{yao2022image}.  Herein, PCL designs three pre-training tasks to enhance the fine-grained alignment between image differences and captions. The authors of PCL pre-train the model with 8K warm-up steps and 250K iterations in total. In contrast to them, the other compared methods are trained in an end-to-end manner. Therefore, we compare PCL with and without pre-training for a fair comparison. 
The results are shown in Table \ref{clevr-total_com} - \ref{type_meteor}.

In Table \ref{clevr-total_com}, we can observe that 1) NCT achieves superior results on most metrics; 2) note that MCCFormers-D is also based on transformer and identifies change based on feature similarity. There are two major differences between it and ours. First, it implements individual feature matching between two sets of features. Instead, our NCT aims to compare two images at neighborhood level to capture contrastive properties between them, which helps perceive fine-grained change while being immune to viewpoint change. Second, different from it based on the standard transformer for caption generation, we exploit syntax dependencies to calibrate decoder, which helps better understand complex syntax structure of change descriptions. 3) Compared with SGCC, the proposed NCT is a little lower on the metrics of METEOR and ROUGE-L. Our conjecture is that SGCC exploits more visual modalities than ours, such as semantic attributes extracted by Yolov4 \cite{bochkovskiy2020yolov4} and image depth maps that are computed by Monodepth2 \cite{godard2019digging}. In contrast, our NCT surpasses SGCC on the other metrics by a large margin.
4) Compared with PCL, NCT surpasses it without pre-training by a large margin. For PCL with pre-training, NCT also outperforms it on the three metrics. For CIDEr, NCT is a little lower. Our conjecture is that PCL leverages three pre-training tasks (with 8K warm-up steps and 250K iterations in total) to augment the model. 

In Table \ref{clevr-sc-com}, it is noted that NCT outperforms the state-of-the-art methods on every metrics, especially improving CIDEr score by a large margin. In Table \ref{type_cider} - \ref{type_meteor}, we compare NCT with state-of-the-art methods under the specific change types using the metrics of CIDEr, SPICE and METEOR. Especially, CIDEr and SPICE are especially designed for evaluating captioning performance.  The results show that our NCT achieves the superior results over the state-of-the-art methods  in almost every category. This shows that our method has a good generalization ability under different change types.

In a word, compared to the state-of-the-art methods in different situations, the proposed NCT achieves the encouraging performance. This superiority results from that 1) the neighborhood feature aggregating and common feature distilling help learn reliable contrastive features and resist irrelevant viewpoint changes; 2) the syntax dependencies can solve the problem of structure ambiguity in change descriptions.

\subsubsection{Results on the CLEVR-DC Dataset}
The experiment is also carried out on a newly released synthetic dataset (ICCV'21) with extreme viewpoint changes. We compare with six state-of-the-art methods: DUDA/DUDA+CC \cite{park2019robust}, M-VAM/M-VAM+CC \cite{shi2020finding}, and VA/VACC\cite{kim2021agnostic}.  

The comparison results are shown in Table \ref{t_dc}. We find that NCT outperforms the state-of-the-art methods on most metrics by a large margin. This validates that our method has a good robustness in any viewpoint change. This mainly benefits from the fact of capturing contrastive information between a pair of images at neighborhood level, because under viewpoint changes, the object relations are stable within/between local neighborhoods of no change.

\begin{table}[t]
  \centering
  \caption{Comparison with the state-of-the-art methods on CLEVR-DC.}
    \begin{tabular}{c|c|c|c|c|c|}
    \hline
    Method & B     & M     & R     & C     & S  \\
    \hline
    DUDA  & \cellcolor[rgb]{ .855,  .91,  .973}40.3 & \cellcolor[rgb]{ .851,  .882,  .949}27.1 & -     & 56.7  & \cellcolor[rgb]{ .706,  .776,  .906}16.1  \\
    DUDA + CC & \cellcolor[rgb]{ .851,  .882,  .949}41.7 & \cellcolor[rgb]{ .851,  .882,  .949}27.5 & -     & \cellcolor[rgb]{ .851,  .882,  .949}62.0 & \cellcolor[rgb]{ .706,  .776,  .906}16.4 \\
    M-VAM & \cellcolor[rgb]{ .851,  .882,  .949}40.9 & \cellcolor[rgb]{ .851,  .882,  .949}27.1 & -     & \cellcolor[rgb]{ .855,  .91,  .973}60.1 & \cellcolor[rgb]{ .851,  .882,  .949}15.8 \\
    M-VAM+CC & \cellcolor[rgb]{ .851,  .882,  .949}41.0 & \cellcolor[rgb]{ .851,  .882,  .949}27.2 & -     & \cellcolor[rgb]{ .851,  .882,  .949}62.0 & \cellcolor[rgb]{ .706,  .776,  .906}16.4 \\
    VA    & \cellcolor[rgb]{ .706,  .776,  .906}44.5 & \cellcolor[rgb]{ .706,  .776,  .906}29.2 & -     & \cellcolor[rgb]{ .706,  .776,  .906}70.0 & \cellcolor[rgb]{ .557,  .663,  .859}17.1 \\
    VACC  & \cellcolor[rgb]{ .706,  .776,  .906}45.0 & \cellcolor[rgb]{ .706,  .776,  .906}29.3 & -     & \cellcolor[rgb]{ .706,  .776,  .906}71.7 & \cellcolor[rgb]{ .557,  .663,  .859}\textbf{17.6}  \\
    \hline
    NCT  & \cellcolor[rgb]{ .557,  .663,  .859}\textbf{47.5} & \cellcolor[rgb]{ .557,  .663,  .859}\textbf{32.5} & \cellcolor[rgb]{ .557,  .663,  .859}\textbf{65.1} & \cellcolor[rgb]{ .557,  .663,  .859}\textbf{76.9} & \cellcolor[rgb]{ .851,  .882,  .949}15.6  \\
\cline{1-5}    \end{tabular}%
  \label{t_dc}%
\end{table}%

\begin{table}[htbp]
  \centering
  \caption{Comparison with the state-of-the-art methods on Image Editing Request Dataset.}
    \begin{tabular}{c|c|c|c|c|c}
    \hline
    Method & B     & M     & R     & C     & S  \\
    \hline
    multi-head att & \cellcolor[rgb]{ 1,  .902,  .6}6.1 & \cellcolor[rgb]{ 1,  .949,  .8}11.8 & \cellcolor[rgb]{ 1,  .949,  .8}35.1 & \cellcolor[rgb]{ 1,  .949,  .89}22.8 & -  \\
    static rel-att & \cellcolor[rgb]{ 1,  .949,  .8}5.8 & \cellcolor[rgb]{ 1,  .902,  .6}12.6 & \cellcolor[rgb]{ 1,  .949,  .8}35.5 & \cellcolor[rgb]{ 1,  .949,  .8}20.7 & - \\
    dynamic rel-att & \cellcolor[rgb]{ 1,  .902,  .6}6.7 & \cellcolor[rgb]{ 1,  .902,  .6}12.8 & \cellcolor[rgb]{ 1,  .902,  .6}37.5 & \cellcolor[rgb]{ 1,  .902,  .6}26.4 & -  \\
    \hline
    NCT  & \cellcolor[rgb]{ 1,  .851,  .4}\textbf{8.1} & \cellcolor[rgb]{ 1,  .851,  .4}\textbf{15.0} & \cellcolor[rgb]{ 1,  .851,  .4}\textbf{38.8} & \cellcolor[rgb]{ 1,  .851,  .4}\textbf{34.2} & \cellcolor[rgb]{ 1,  .851,  .4}\textbf{12.7}  \\ \hline
    \end{tabular}%
  \label{t_edit}%
\end{table}%

\subsubsection{Results on the Image Editing Request Dataset}
We conduct the experiment on another challenging dataset, Image Editing Request. The changed objects in this dataset are usually vague and inconspicuous.  We  compare NCT with three state-of-the-art methods reported by Tan \emph{et al.} \cite{tan2019expressing}: multi-head att, static rel-att, and dynamic rel-att. 

Table \ref{t_edit} shows that NCT outperforms the state-of-the-art methods by a large margin. This indicates that the proposed method can accurately describe which part of the ``source'' image has been edited by capturing neighborhood contrastive features and achieving syntax disambiguity based on explicit dependencies between words.  

In short, experiments on the above three datasets show that our method has a good generalization of change localization and description on different change scenarios.
\begin{table}[t]
  \centering
  \caption{Ablation studies based on total performance on CLEVR-Change}
    \begin{tabular}{c|c|c|c|c|c}
    \hline
    \multicolumn{1}{c|}{Method} & B     & M     & R     & C     & S  \\
    \hline
    Diff-sub & 53.3  & 38.8  & 72.1  & 119.7 & \cellcolor[rgb]{ .949,  .949,  .949}31.8  \\
    NFA   & \cellcolor[rgb]{ .929,  .929,  .929}54.3 & \cellcolor[rgb]{ .929,  .929,  .929}39.7 & \cellcolor[rgb]{ .929,  .929,  .929}73.1 & \cellcolor[rgb]{ .949,  .949,  .949}121.9 & \cellcolor[rgb]{ .929,  .929,  .929}32.0 \\
    CFD   & \cellcolor[rgb]{ .929,  .929,  .929}54.1 & \cellcolor[rgb]{ .929,  .929,  .929}39.6 & \cellcolor[rgb]{ .929,  .929,  .929}73.1 & \cellcolor[rgb]{ .929,  .929,  .929}122.8 & \cellcolor[rgb]{ .929,  .929,  .929}32.2 \\
    
    ST    & \cellcolor[rgb]{ .949,  .949,  .949}53.7 & \cellcolor[rgb]{ .929,  .929,  .929}39.4 & \cellcolor[rgb]{ .949,  .949,  .949}72.7 & \cellcolor[rgb]{ .949,  .949,  .949}120.7 & \cellcolor[rgb]{ .949,  .949,  .949}31.9  \\
    NCT w/o S   & \cellcolor[rgb]{ .859,  .859,  .859}54.6 & \cellcolor[rgb]{ .859,  .859,  .859}40.0 & \cellcolor[rgb]{ .929,  .929,  .929}73.6 & \cellcolor[rgb]{ .859,  .859,  .859}123.4 & \cellcolor[rgb]{ .859,  .859,  .859}32.5 \\
    \hline
    NCT  & \cellcolor[rgb]{ .788,  .788,  .788}\textbf{55.1} & \cellcolor[rgb]{ .788,  .788,  .788}\textbf{40.2} & \cellcolor[rgb]{ .788,  .788,  .788}\textbf{73.8} & \cellcolor[rgb]{ .788,  .788,  .788}\textbf{124.1} & \cellcolor[rgb]{ .788,  .788,  .788}\textbf{32.9}  \\
    \hline
    \end{tabular}%
  \label{ablation_total}%
\end{table}%

\subsection{Ablation Studies}

We carry out ablation studies to validate the effectiveness of each proposed module and the full model. (1) Diff-sub is a transformer-based baseline model which computes difference features by direct subtraction. Specifically, it first directly subtracts two image features to obtain the difference representation. Then, it uses the spatial attention mechanism to select the most relevant features on each image based on the subtracted representation. Finally, the specifically changed features are fed into a standard transformer decoder for caption generation.	 
(2) NFA performs neighborhood feature aggregating before subtraction. (3) CFD distills common features to construct contrastive features, instead of using direct subtraction. (4) ST refers to syntax-aware transformer decoder that uses syntax dependency relation to augment the baseline model.  (5) NCT w/o S is the neighborhood contrastive transformer without syntax dependency.  (6) NCT is the proposed full model: neighborhood contrastive transformer with syntax dependency.

The ablation studies are based on the total performance on CLEVR-Change, and the results are shown in Table~\ref{ablation_total}. We observe that 1) compared to the baseline model, each module and the full model achieve consistent improvements; 2) the performance of NFA and CFD are close, and better performance is achieved by their combination; 3) only augmenting the baseline model with syntax dependency, the improvement is slight, and the best performance is achieved through combining it with NCT. The above observations indicate that 1) the effectiveness of each proposed module and the full model; 2) each module not only plays its unique role, but also supplements the other; 3) only if the model learns effective contrastive features by neighborhood feature aggregating and common feature distilling, the syntax-aware decoder would use these features to yield correct sentences. 

\subsection{Evaluating the Model's Robustness under Various Degrees of Viewpoint Changes}
\begin{figure}[t]
\includegraphics[width=0.5\textwidth]{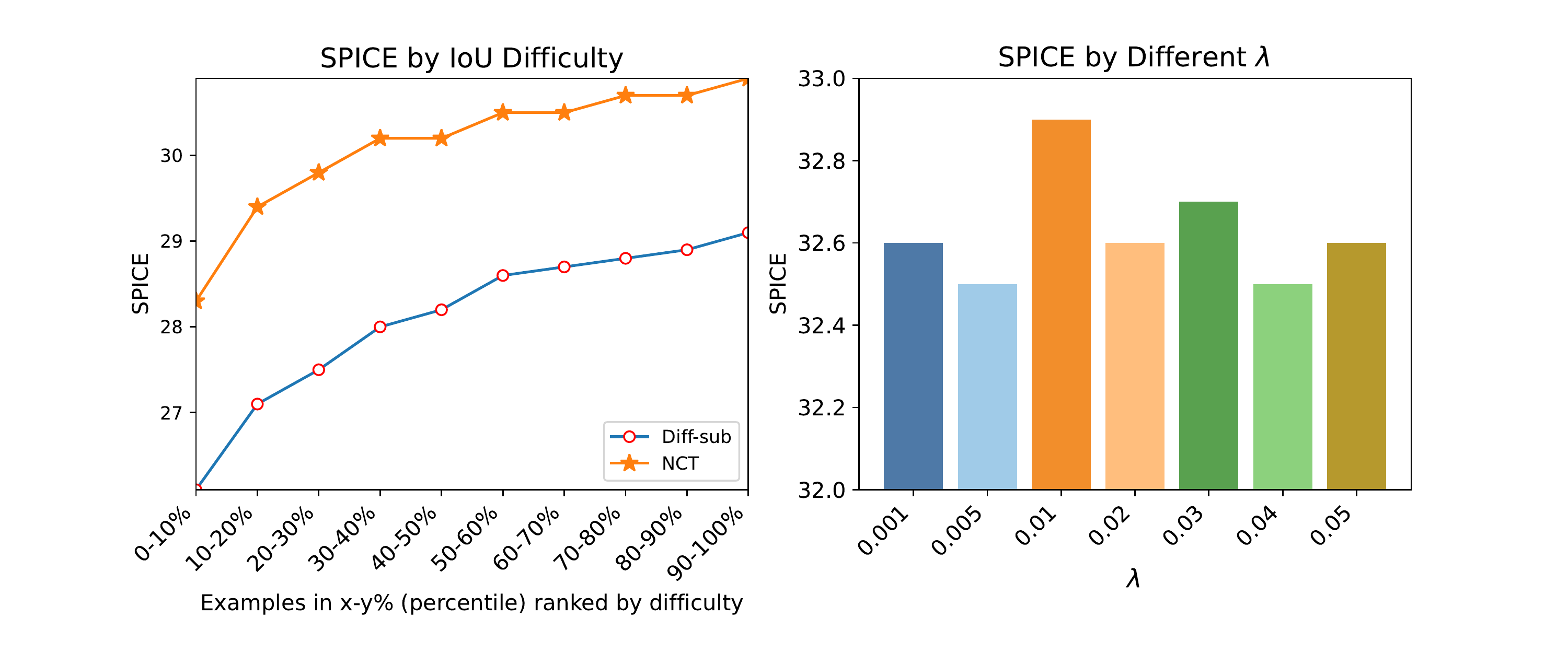} 
\caption{Left sub-figure is the visualization of captioning performance (SPICE) that is breakdown by viewpoint change (measured by IoU); right sub-figure is the effects of the trade-off parameter $\lambda$ on CLEVR-Change. }
\label{iou}
\end{figure}

To evaluate the robustness of NCT under different degrees of viewpoint changes, following the pioneer work \cite{park2019robust}, we compute the IoU of the bounding boxes of the objects (except the changed object) across the two images, where the lower IoU refers to higher difficulty.
Herein, we employ SPICE to evaluate the sentences generated by Diff-sub (baseline model) and NCT with respect to different IoU of the object’s bounding boxes.
The results are shown in the left sub-figure of Fig. \ref{iou}.
It is noted that NCT consistently outperforms the baseline by a large margin. This indicates that the proposed method can identify reliable change and handle the varying degrees of  viewpoint changes.

\subsection{Study on the Trade-off Parameter $\lambda$}
\begin{figure}[t]
\includegraphics[width=0.47\textwidth]{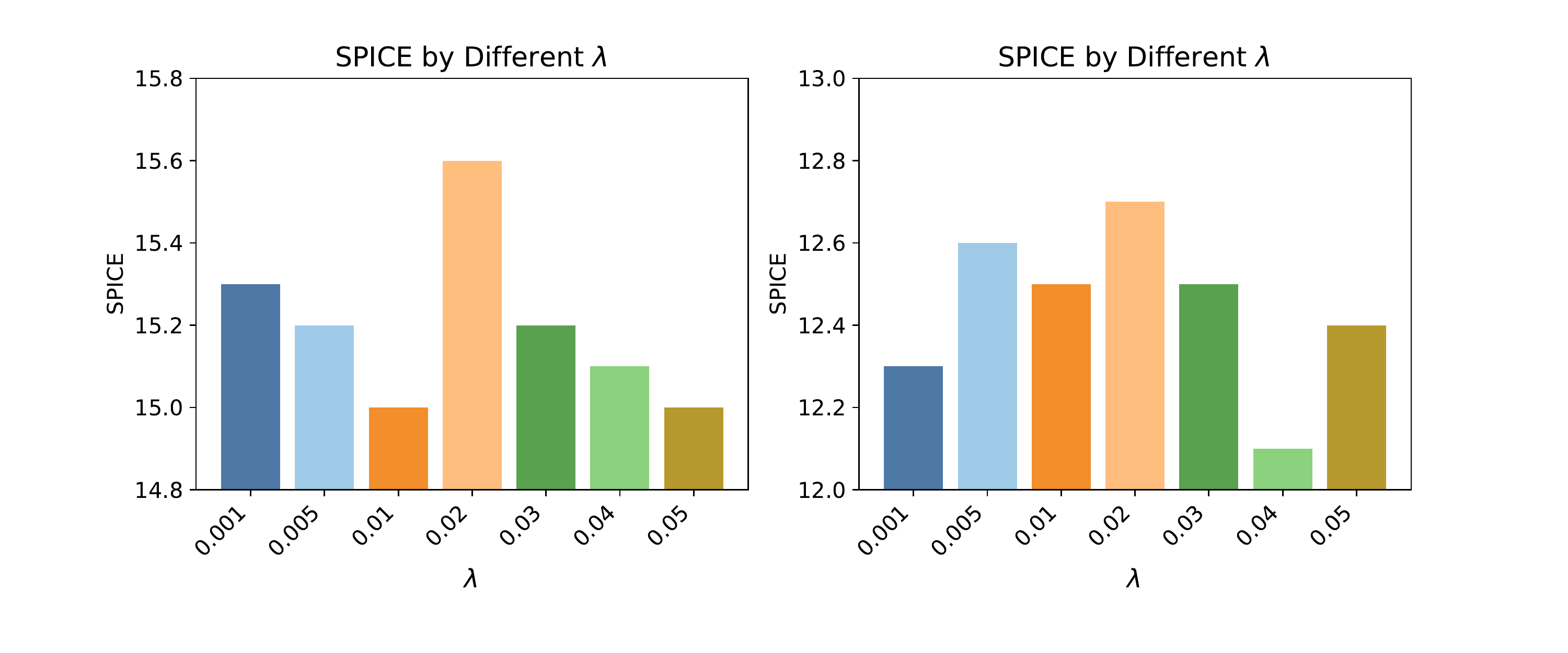} 
\caption{The effects of the trade-off parameter $\lambda$ on CLEVR-DC (left sub-figure) and Image Editing Request (right sub-figure). }
\label{spice}
\end{figure}

In this section, we discuss the effect of the trade-off
parameter $\lambda$ in Eq. (\ref{cross-entropy}) on CLEVR-Change, CLEVR-DC and Image Editing Request. This parameter is to balance the contributions from the caption generator and dependency predictor. On CLEVR-Change, with different values, the obtained SPICE scores are shown in the right sub-figure of Fig. \ref{iou}. We find that as the values of $\lambda$ increasing or decreasing, the performance of NCT changes. This is mainly because the whole model will focus much on one part but ignore the supervision signal from the other. Based on this, we empirically set $\lambda$ to 0.01. In addition, for other two datasets, CLEVR-DC and Image Editing Request, the results are shown in Fig. \ref{spice}.
With different values, the obtained SPICE scores on CLEVR-DC in the left sub-figure, and on Image Editing Request in the right sub-figure. It is noted that similar to the experimental results on CLEVR-Change, as the values of $\lambda$ increasing or decreasing, the performance of NCT changes. On the both datasets, the better value is 0.02. The above analysis shows that the value of this trade-off parameter $\lambda$ is close on different datasets, which validates that the proposed method has a good robustness on different change scenarios.

\begin{figure*}[t]
\includegraphics[width=1\textwidth]{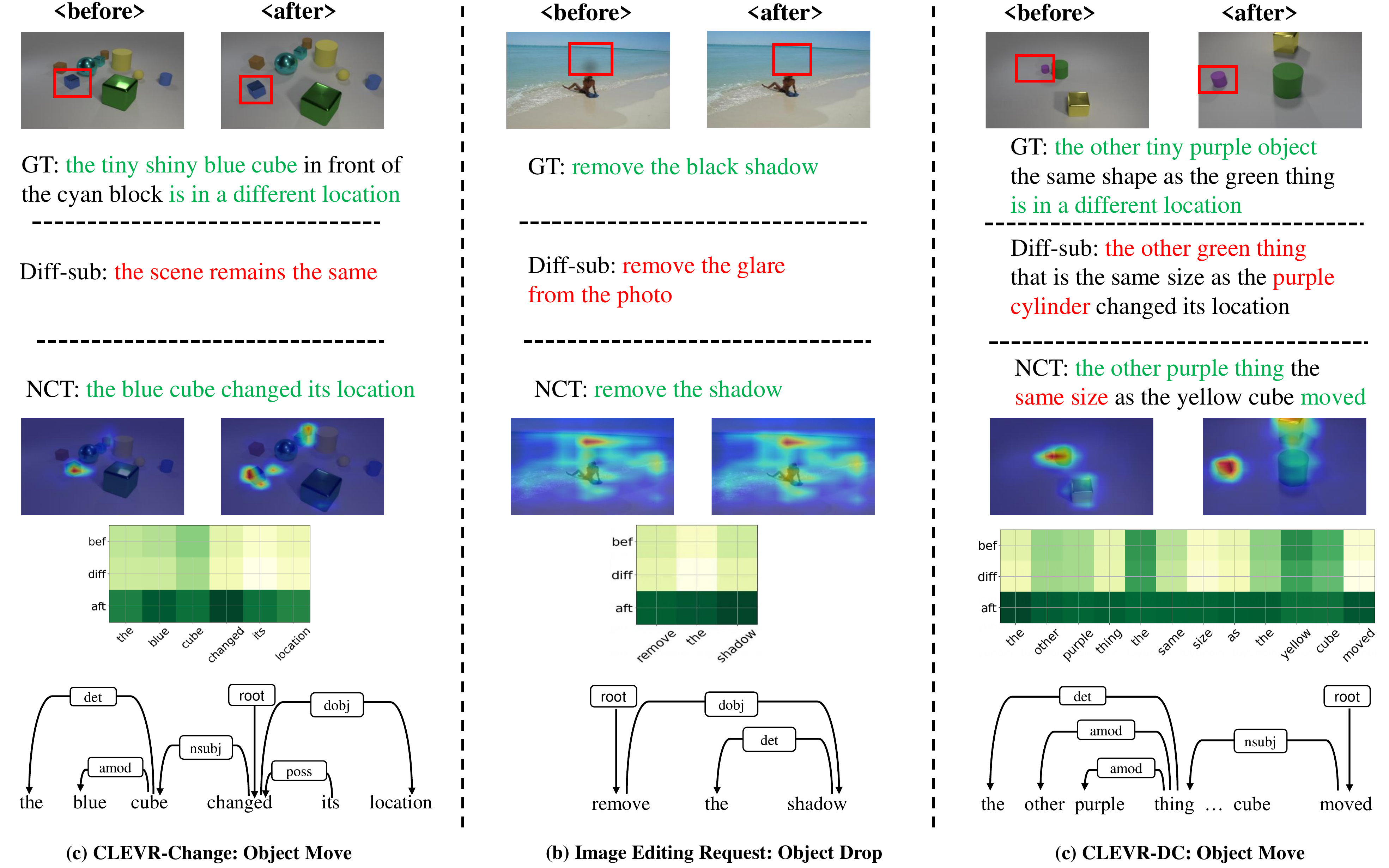} 
\caption{Qualitative examples on CLEVR-Change, Image Editing Request, and CLEVR-DC. For each example, we report the captions generated by Diff-sub and NCT along with the ground-truth (GT) captions.  Correct and incorrect parts of the captions are in green and red, respectively.  We visualize the results of change localization on the ``before'' and ``after'' images, and illustrate the heat map visualization to show the semantic alignment between changed features and corresponding words. We visualize the predicted dependencies of each example. The ground-truth changes are shown in red boxes.}

\label{q_trans}
\end{figure*}

\subsection{Study on the Parameter of Neighborhood Range $r$}
 In this section, we will  analyze the effect of neighborhood range $r$.  Herein, we set it as 3 and 5, respectively. Note that as this value is larger than 5, the computation cost increases sharply and is more than one RTX 3090 GPU. With different values, the captioning performance and parameter number are shown in Table \ref{range}.

We find that there is no obvious performance increase as we enlarge neighborhood range, and the results on most metrics even decrease. Our conjecture is that the model only needs the closet referents to guide where the changed object is, so the neighborhood range of 3$\times $3 is suitable. Based on this, we empirically set $r$ to 3 on the three datasets.

\begin{table}[htbp]
  \centering
  \caption{Study the effects of the parameter of neighborhood range $r$ on the three datasets, where CC, CD, and IER are short for CLEVR-Change, CLEVR-DC, and Image Editing Request.}
    \begin{tabular}{c|c|c|c|c|c|c}
    \hline
    \multicolumn{1}{c|}{$r$ } & Set & Params & B     & M     & C     & S  \\
    \hline
    3 $\times$ 3 & CC    & 26.65M & \textbf{55.1} & \textbf{40.2} & \multicolumn{1}{c|}{124.1} & \textbf{32.9}  \\
    5 $\times$ 5 & CC    & 26.74M & 54.9  & 39.9  & \textbf{124.8} & 32.7  \\
    \hline
    3 $\times$ 3 & CD    & 26.70M & \textbf{47.5} & \textbf{32.5} & \textbf{76.9} & \textbf{15.6}  \\
    5 $\times$ 5 & CD    & 26.79M & 44.4  & 31.3  & 71.1  & 14.5 \\ \hline
    \multicolumn{1}{c|}{3 $\times$ 3} & IER   & 34.07 M & 8.1   & \textbf{15.0} & 34.2  & \textbf{12.7} \\
    5 $\times$ 5 & IER   & 34.17M & \textbf{9.5} & 14.7  & \textbf{36.9} & 11.9  \\
    \hline
    \end{tabular}%
  \label{range}%
\end{table}%

\subsection{Qualitative Analysis}
To evaluate the overall performance of NCT about change localization and caption generation, we conduct qualitative analysis on CLEVR-Change, Image Editing Request, and CLEVR-DC, as shown in Fig. \ref{q_trans}. For each image pair, we report the captions yielded by the baseline model of Diff-sub and our NCT along with the ground truth (green words). To evaluate the accuracy of changed objects, we also visualize the changed results based on the attention weights of change detection. 
For the first example, the object movement is slight, which makes Diff-sub misjudge that there is nothing changed.   For the second example, the removed object is too faint to notice. In this case, Diff-sub  directly subtracts two images to compute change features, which wrongly judges the ``shadow'' as ``glare'' and fails to generate the accurate sentence.
For the third example, extreme viewpoint change results in pseudo movements of all objects, which makes Diff-sub misidentify really changed object. Besides, another possible reason of this failure is that the syntax structure in the ground-truth sentence is complex. That is, the referent ``green thing'' is closer to changed type than ``purple thing'', which might make Diff-sub misjudge the changed object as the ``green thing''. In contrast to Diff-sub, the proposed NCT can accurately localize and describe changed objects. This mainly benefits from that 1) the neighboring feature aggregating helps the model identify the real change while being immune to viewpoint change; 2) the common feature distilling can effectively summarize common properties of the image pair and extract differentiating features from each image, so as to construct constrictive features between them; 3) introducing dependency relations between words helps solve syntax ambiguity in sentences and  understand their complex syntax structure. For instance, in the third example, 
 our NCT can predict the directed link between really changed object ``purple thing'' and changed type ``moved'', so as to identify the really changed object. 
 
 In addition, we find that in the third example, NCT predicts that the tiny purple object is the same size as the yellow cube. The possible reasons for this misunderstanding are that 1) NCT compares their size mainly based on the ``after'' image. 2) The change types in this dataset do not include size change. In this case, the model has a limited ability to accurately compare the size between two objects. Therefore, further exploration to identify the changes of objects' size is warranted in future research.  More qualitative examples  are shown in the supplementary material.

\begin{figure*}[htbp]
\centering
\includegraphics[width=0.9\textwidth]{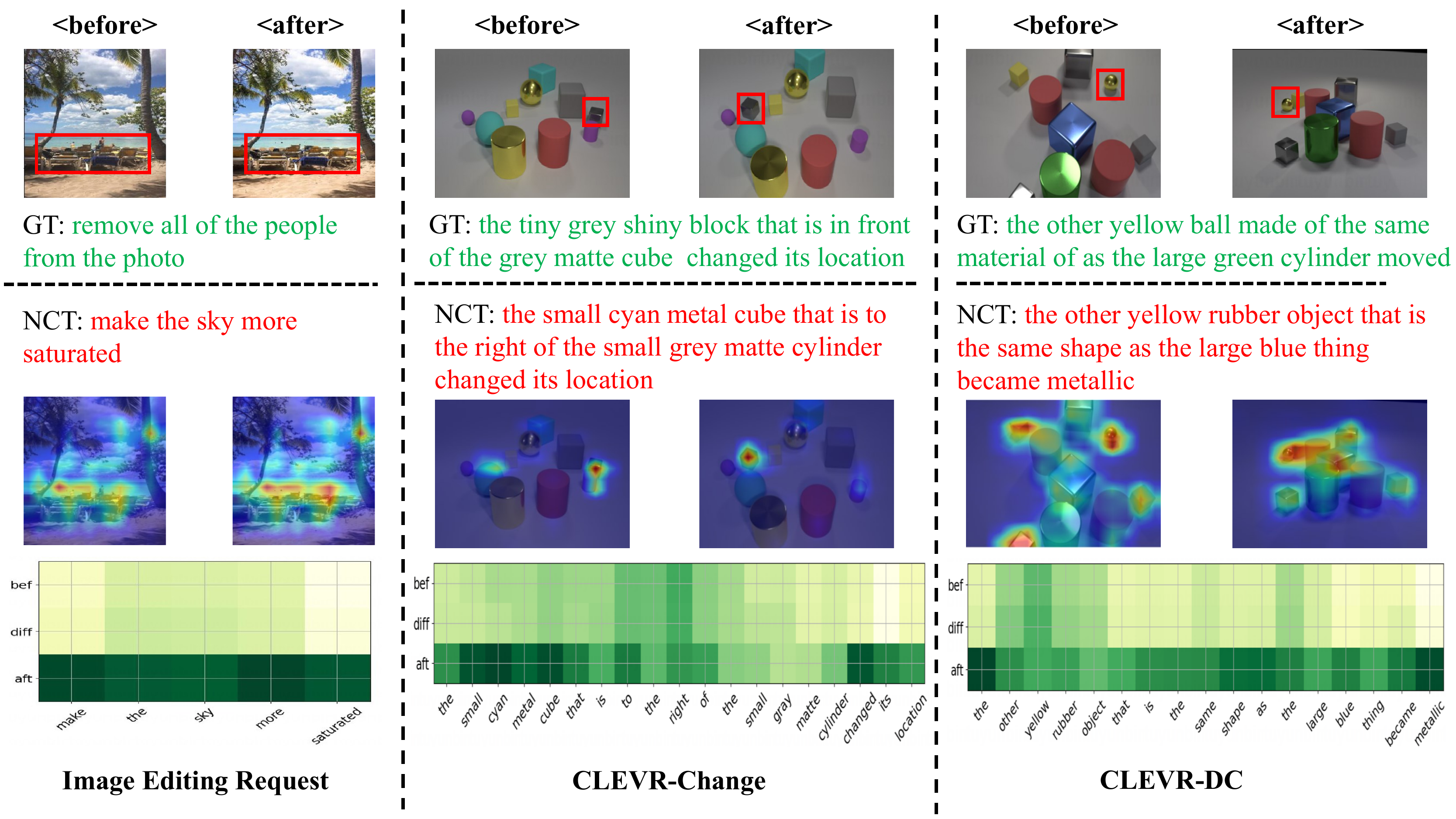} 
\caption{Failure examples obtained by NCT on the test split of Image Editing Request, CLEVR-Change, and CLEVR-DC. }
\label{failure}
\end{figure*}

\begin{figure*}[t]
\centering
\includegraphics[width=0.9\textwidth]{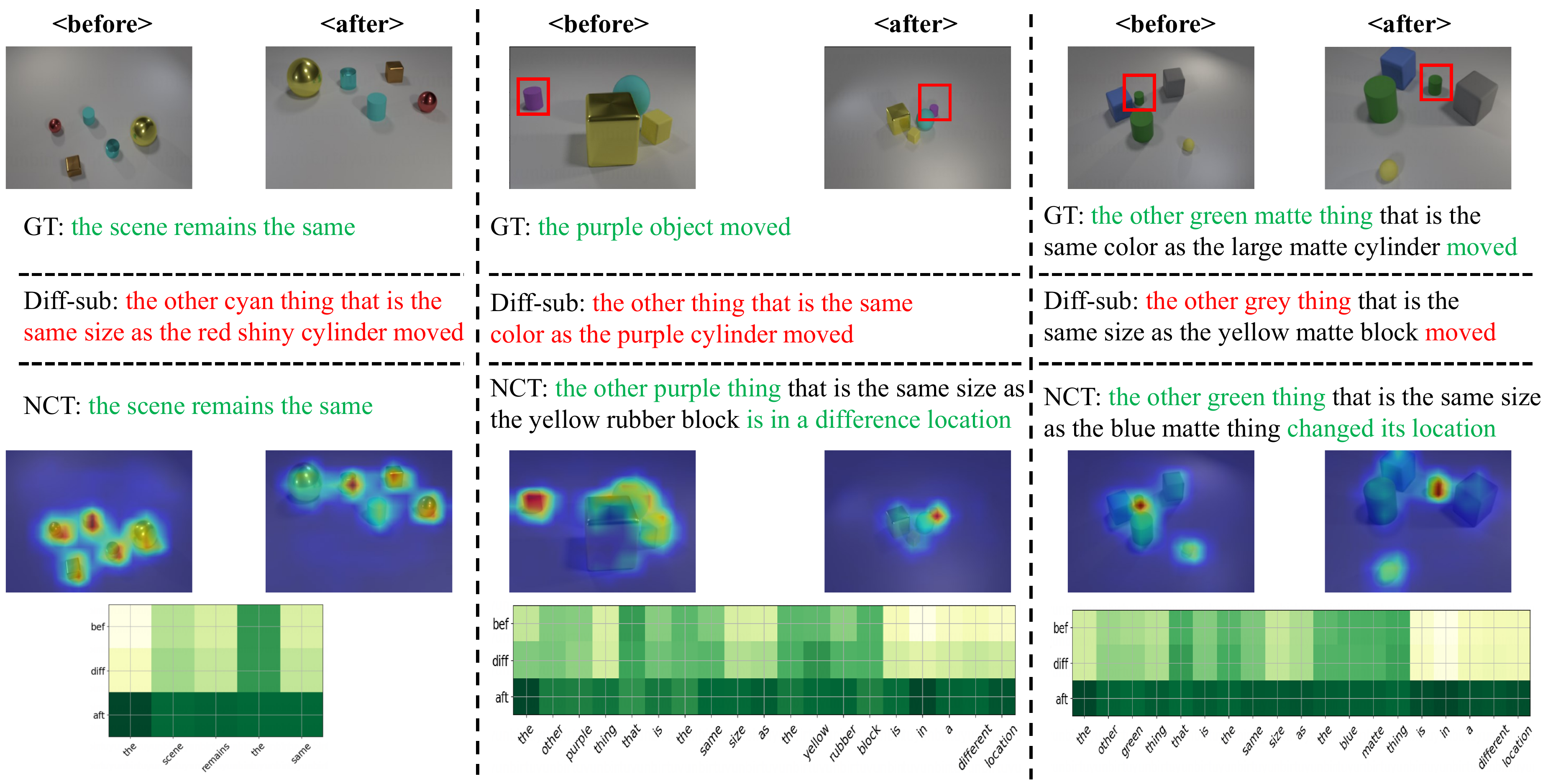} 
\caption{Qualitative examples on the test split of CLEVR-DC.} 
\label{dc}
\end{figure*}

\begin{figure*}[htbp]
\centering
\includegraphics[width=0.88\textwidth]{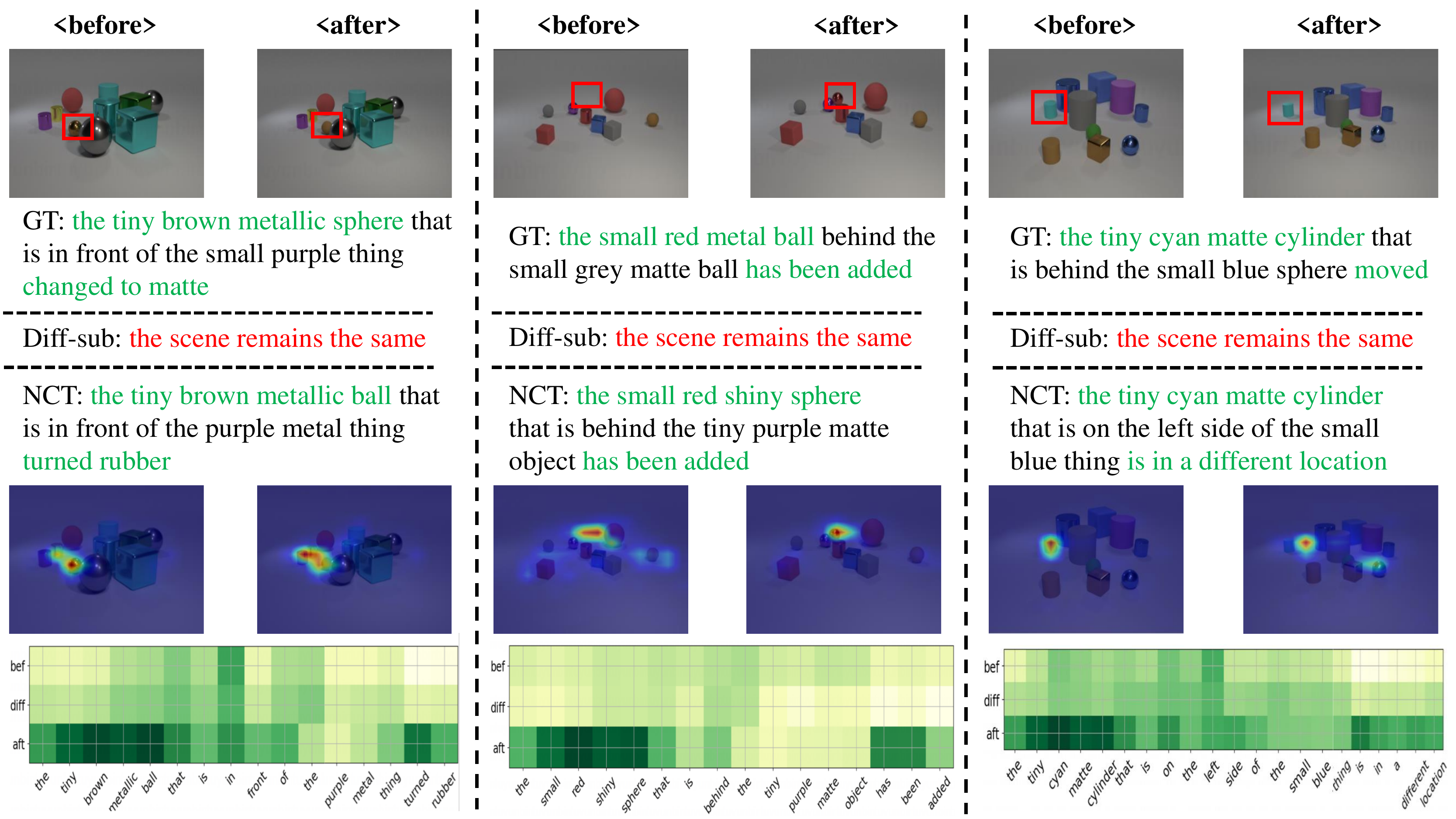} 
\caption{Qualitative examples on the test split of CLEVR-Change. }
\label{change}
\end{figure*}

\begin{figure*}[htpb]
\centering
\includegraphics[width=0.9\textwidth]{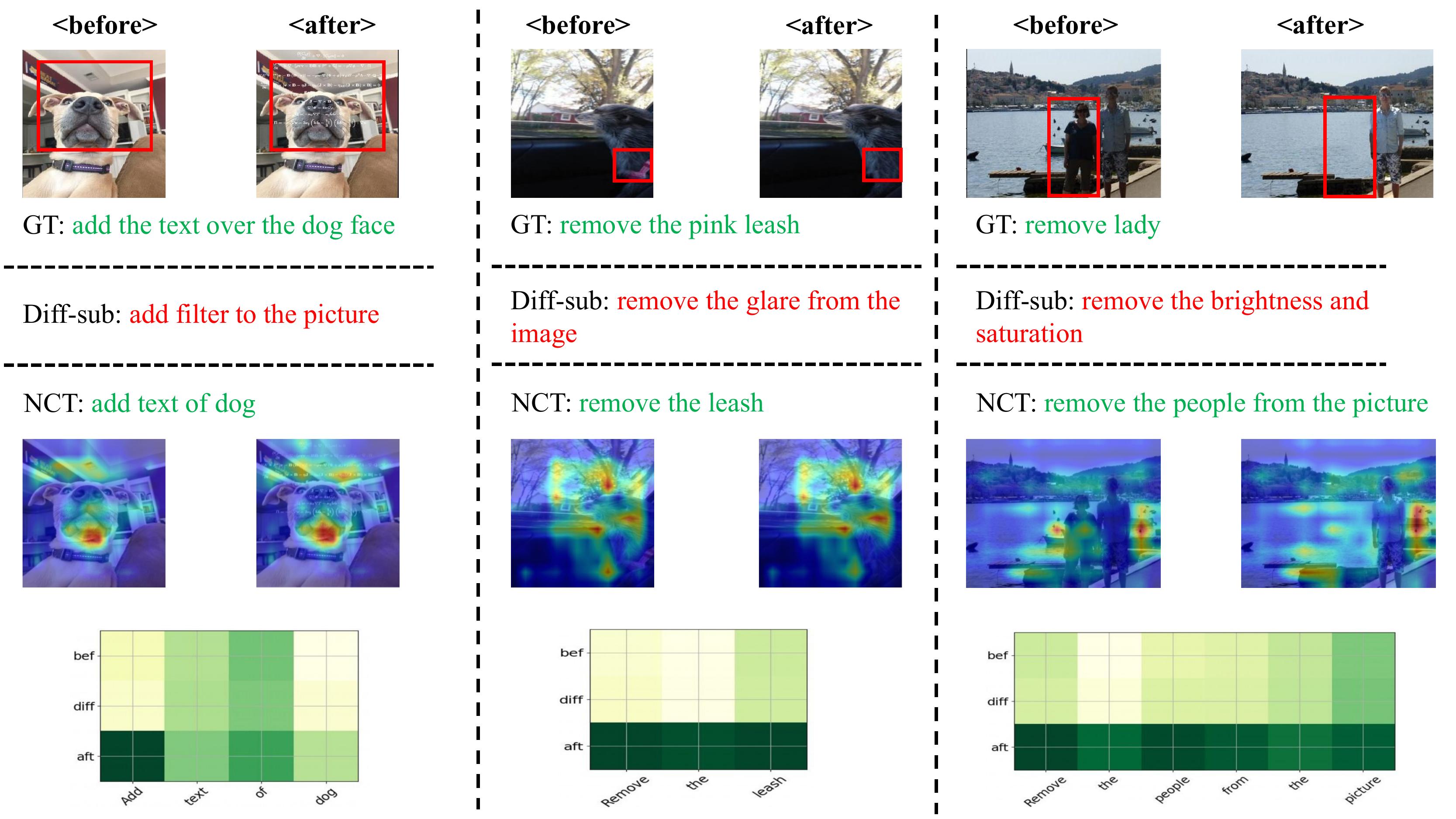} 
\caption{Qualitative examples on the test split of Image Editing Request. }
\label{edit}
\end{figure*}

\subsection{Discussion}
Fig. \ref{failure} illustrates the failure cases obtained by NCT on the three datasets with different change scenarios. For all the change scenes, we can observe that 
the proposed NCT successfully localizes the changed objects. However, it fails to describe them in accurate sentences. For the failure cases, our conjecture is that the visual signal of change appears in a inconspicuous region with weak feature in each example. This makes it overwhelmed by   most unchanged objects. As such, the decoder cannot receive sufficient visual information for caption generation. In our opinion, there are two possible solutions for this challenge. One solution is to exploit other visual modalities to augment grid features, such as semantic segmentation features which can capture more fine-grained visual information \cite{wu2022difnet}, so as to enhance the feature representation for these objects with weak change signals. The other solution is to take advantage of  the paradigm of pre-training to fine-tuning, such as  fine-tuning the visual features on change captioning datasets. Specifically, we can exploit a pre-trained feature extractor that coordinates with our framework (\emph{e.g.}, Vision Transformer \cite{DBLP:conf/iclr/DosovitskiyB0WZ21}). Then, we do not freeze its parameters and jointly train it with the proposed NCT. In this way, the training loss can be propagated back to the feature extractor, so as to enhance the representation ability of image features. We will try these strategies in the subsequent work. In addition, we notice that the visualized attention weights of change localization are with noises on CLEVR-DC. we will try to address the problem of extreme viewpoint changes from the perspective of leveraging 3D knowledge in the future.

\section{Conclusion}
\label{conclusion}
In this paper, we propose a Neighborhood Contrastive Transformer (NCT) to pinpoint and describe the change under different change scenes. In NCT, the neighborhood feature aggregating module can help overcome the influence of viewpoint change, and quickly find the inconspicuous change under the guidance of surrounding conspicuous referents. The common feature distilling module can capture common properties from each image and learn contrastive representation between the image pair. Furthermore, we introduce the explicit dependencies between words to calibrate the decoder of transformer, which helps understand complex syntax structure in change descriptions during training.
Extensive experiments demonstrate that NCT outperforms the state-of-the-art methods by a large margin on the three public datasets with different change scenarios, which also shows that it has a good generalization ability to deal with various change settings.


{\appendix [Implementation Details and More Qualitative Examples on the Three Datasets]

In the appendix, we first provide more implementation details of our method. On the three datasets, we train the model to convergence with 10K iterations in total. Both training and inference are implemented with PyTorch on an RTX 3090 GPU. In the training stage, the used resources on the three datasets are shown in Table \ref{training time}. We can find that our method does not need much resources and training time, so it can be easy reproduced by   other researchers.
\begin{table}[htbp]
  \centering
  \caption{The usage of training time and GPU Memory on the three datasets}

    \begin{tabular}{c|c|c}
   \hline
    \multicolumn{1}{c|}{} & Training Time & GPU Memory \\
    \hline
    CLEVR-Change & 4 hours & 13G \\
    CLEVR-DC & 2 hours & 8G \\
    Image Editing Request & 30 minutes  & 4G \\
 \hline
    \end{tabular}%
  \label{training time}%
\end{table}%

Then, we illustrate more qualitative examples about change localization and caption generation on the three datasets, which are shown in Fig. \ref{dc} - \ref{edit}. For the CLEVR-DC dataset that is to stimulate extreme viewpoint changes, there exist obvious pseudo movements for all the objects in a scene, as shown in Fig. \ref{dc}. This misleads the baseline model of Diff-sub into yielding wrong results. The qualitative examples on CLEVR-Change are shown in Fig. \ref{change}. Since the changed objects are partially occluded or inconspicuous, the baseline model cannot locate these changes and misjudges nothing has changed. Instead, the proposed NCT accurately distinguishes these fine-grained changes from pseudo changes and generates related sentences. It is noted that in} Fig. \ref{dc}, the heat maps in the left-hand side example highlight all the five objects. Our conjecture is that the heat map is generated based on the attention weights of contrastive change localizer (Sec. \ref{localizer}). When nothing has changed, the learned contrastive representation of the image pair would not contain the information of changed object. And the features of background are much weaker than the object features. In this case, the localizer would attend to object features and assign similar attention weight for each object feature, so the visualized heat maps highlight all the five objects.
On Image Editing request from Fig. \ref{edit}  we can observe that  in each example, the change information is so vague that it is hard to find, but our model still locates the changed object, so as to generate the desirable caption compared to the baseline model. The superior results of our method mainly benefits from that 1) the neighborhood feature aggregating helps the model handle irrelevant viewpoint change and locate fine-grained change; 2) the common feature distilling can capture joint information of the image pair and extract differentiating properties from each image, which constructs constrictive features between them; 3) introducing explicit dependency relations between words helps disambiguate complex syntax structure in change sentences during training.

\bibliographystyle{IEEEtran}
\bibliography{IEEEabrv,IEEEfull}

\begin{IEEEbiography}[{\includegraphics[width=1in,height=1.25in,clip,keepaspectratio]{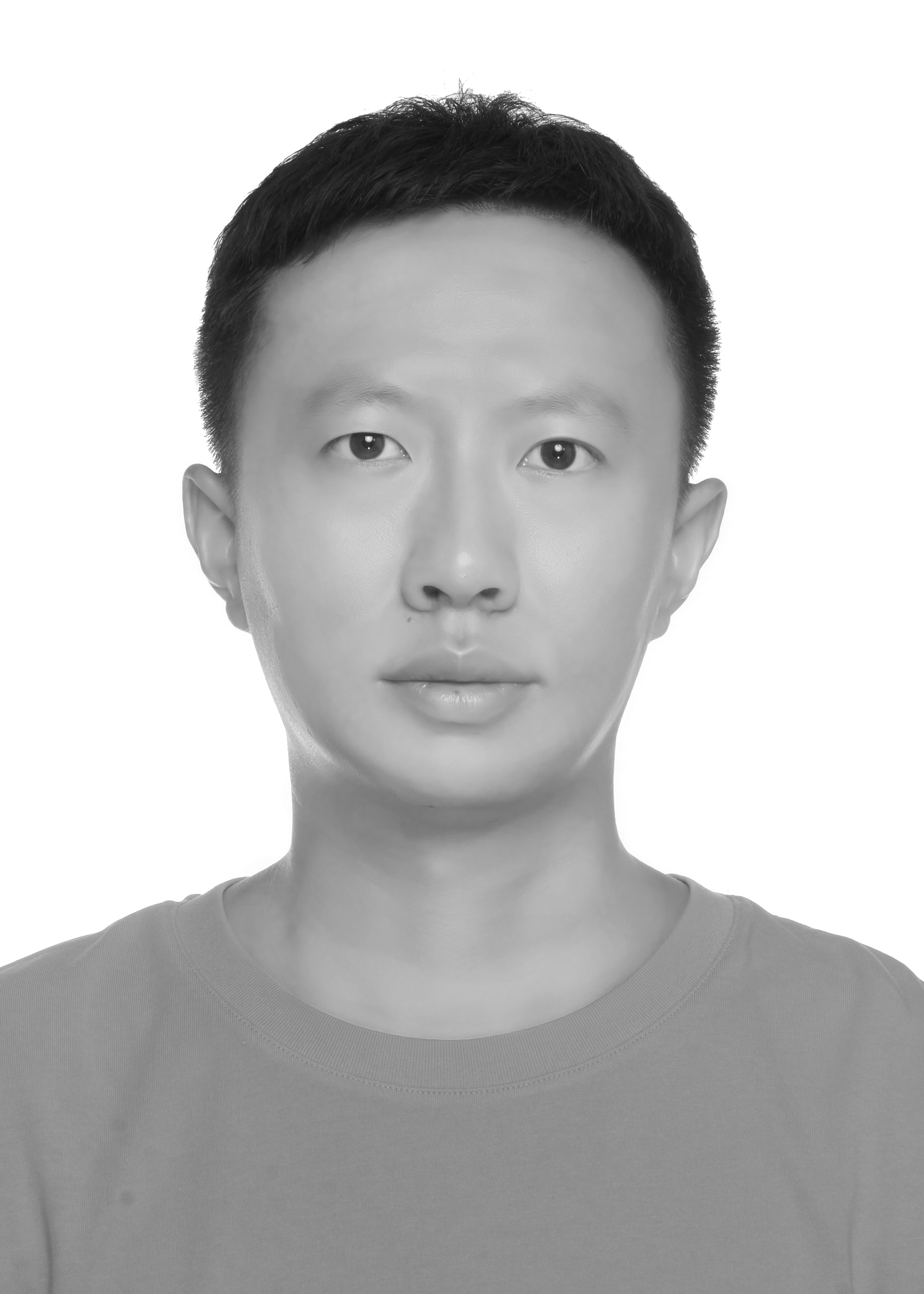}}]{Yunbin Tu} received the B.S. degree in Automation from Hangzhou Dianzi University, and the M.S. degree in Pattern Recognition and Intelligent System from Kunming University of Science and Technology. He is currently pursuing the Ph.D. degree from the School of Computer Science and Technology, University of Chinese Academy of Sciences. His research interests include multimedia content analysis, especially for video and change captioning.

\end{IEEEbiography}

\begin{IEEEbiography}[{\includegraphics[width=1in,height=1.25in,clip,keepaspectratio]{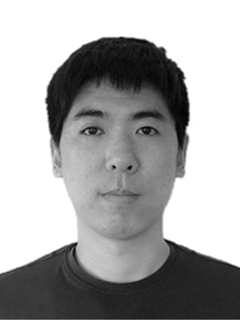}}]{Liang Li} received his B.S. degree from Xi’an Jiaotong University in 2008, and Ph.D. degree from Institute of Computing Technology, Chinese Academy
of Sciences, Beijing, China in 2013. From 2013
to 2015, he held a post-doc position with the Department of Computer and Control Engineering,
University of Chinese Academy of Sciences, Beijing, China. Currently he is serving as the associate
professor at Institute of Computing Technology, Chinese Academy of Sciences. He has also served on a
number of committees of international journals and
conferences. Dr. Li has published over 60 refereed journal/conference papers.
His research interests include multimedia content analysis, computer vision,
and pattern recognition.

\end{IEEEbiography}

\begin{IEEEbiography}[{\includegraphics[width=1in,height=1.25in,clip,keepaspectratio]{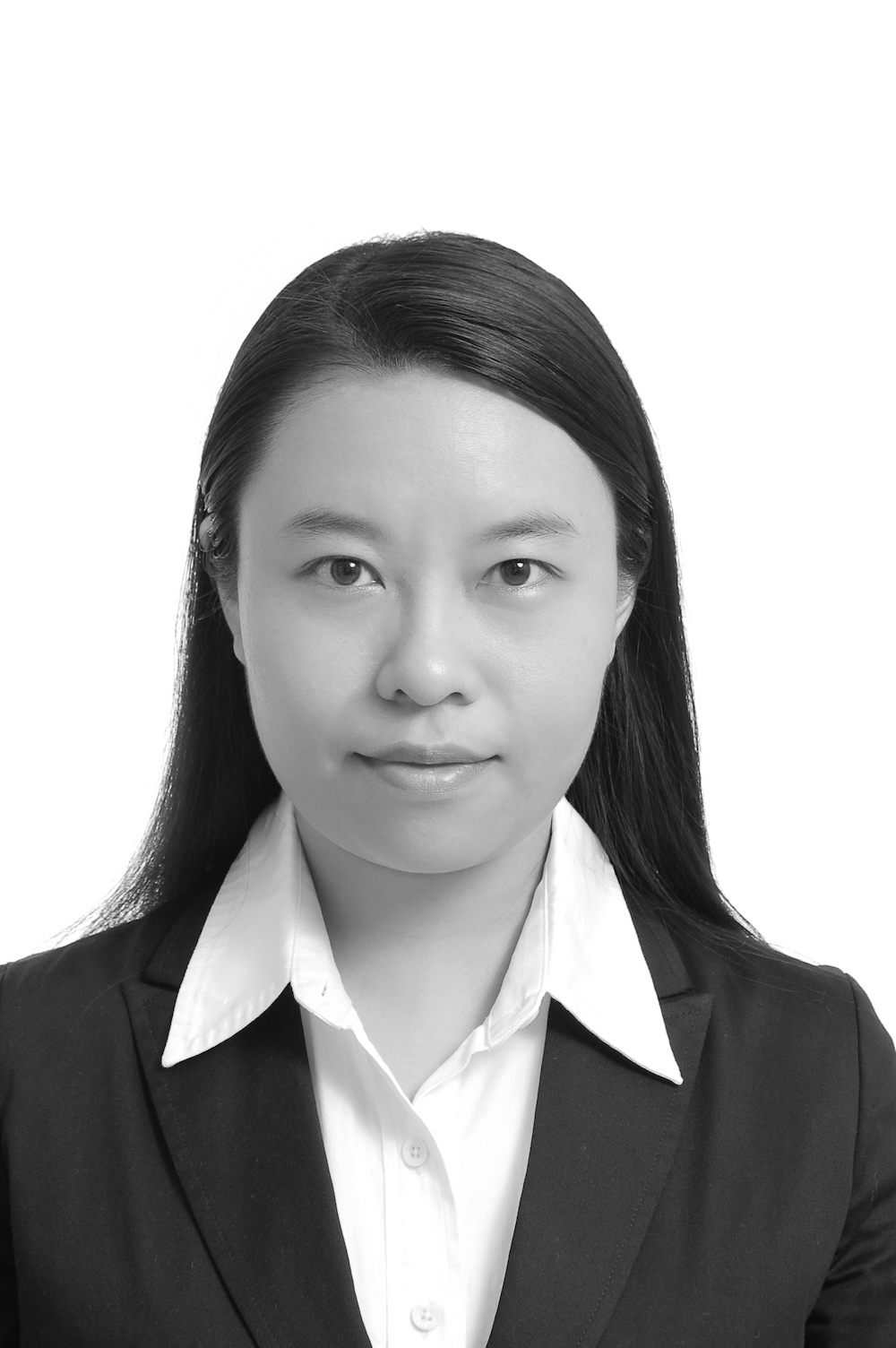}}]{Li Su} received the Ph.D. degree in computer science from the Graduate University of Chinese Academy of Sciences, Beijing, in 2009. She is currently a Professor with the School of Computer Science and Technology, University of Chinese Academy of Sciences. Her research interests include media computing and content analysis.

\end{IEEEbiography}

\begin{IEEEbiography}[{\includegraphics[width=1in,height=1.25in,clip,keepaspectratio]{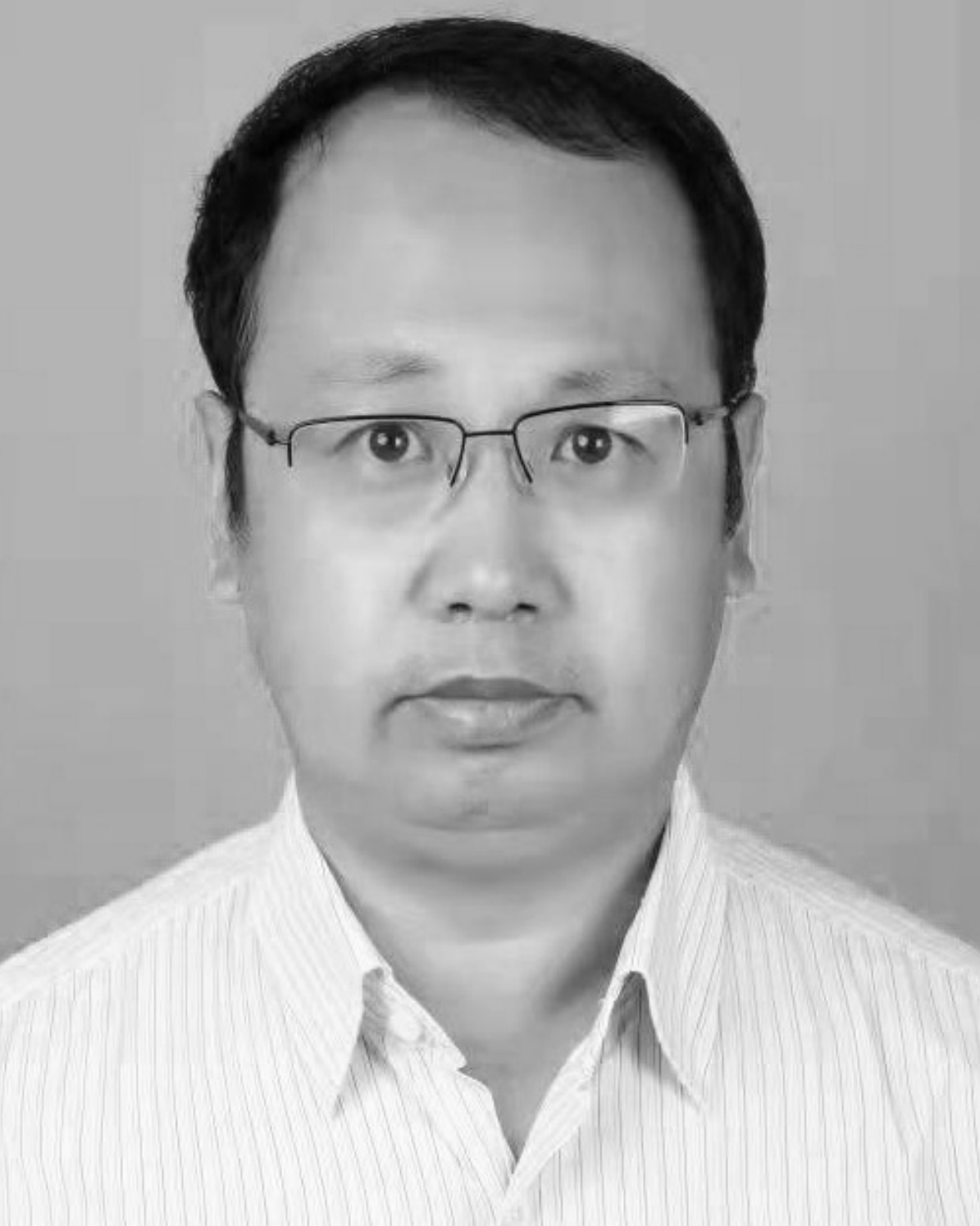}}]{Ke Lu} (Senior Member, IEEE) was born in Ningxia on March 13, 1971. He received the master’s and Ph.D. degrees from the Department of Mathematics and Department of Computer Science, Northwest University, Xi'an, Shanxi, China, in July 1998 and July 2003, respectively.
He worked as a Post-Doctoral Fellow with the Institute of Automation, Chinese Academy of Sciences, Beijing, China, from July 2003 to April 2005. Currently, he is a Distinguished Professor with the University of the Chinese Academy of Sciences, Beijing. He is also a Double-hired Professor with the Pengcheng Laboratory, Shenzhen, Guangdong, China. His current research areas focus on computer vision, 3-D image reconstruction, and computer graphics.
\end{IEEEbiography}

\begin{IEEEbiography}[{\includegraphics[width=1in,height=1.25in,clip,keepaspectratio]{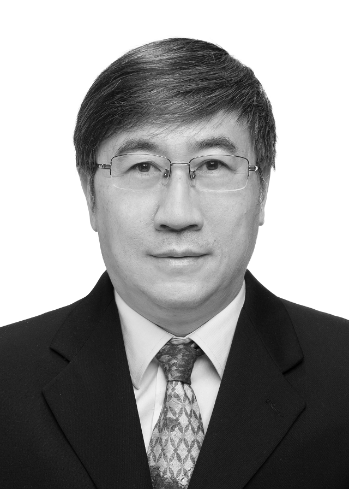}}]{Qingming Huang} received the B.S. degree in computer science and the Ph.D. degree in computer engineering from Harbin Institute of Technology, Harbin,
China, in 1988 and 1994, respectively. He is currently a Chair Professor and the Deputy Dean with the
School of Computer Science and Technology, University of Chinese Academy of Sciences. He has coauthored over 400 academic papers in international
journals, such as IEEE TPAMI, TIP, TKDE, TMM
and TCSVT, and top level international conferences,
including NeurIPS, ACM Multimedia, ICCV, CVPR,
ECCV, VLDB, AAAI and IJCAI. He is a Fellow of IEEE. His current
research interests include multimedia computing, image/video processing,
pattern recognition, and computer vision.
\end{IEEEbiography}

\vfill

\end{document}